\documentclass[10pt,twocolumn,letterpaper]{article}

\usepackage{cvpr}              
\definecolor{cvprblue}{rgb}{0.21,0.49,0.74}
\usepackage[pagebackref,breaklinks,colorlinks,allcolors=cvprblue]{hyperref}
\usepackage{multirow}
\usepackage{amsmath}
\usepackage{wrapfig}
\usepackage{booktabs}
\usepackage{xcolor}
\usepackage{adjustbox}
\usepackage{subcaption}
\usepackage{layouts}
\usepackage{lipsum}
\usepackage{url}
\usepackage{makecell}
\usepackage{tabularray}
\usepackage{algorithm,algorithmic}
\usepackage{amssymb}
\usepackage{pifont}
\usepackage{etoolbox}
\usepackage{colortbl}  
\usepackage{xcolor}
\usepackage{xspace}
\usepackage{float}
\usepackage{csquotes}
\usepackage[accsupp]{axessibility}  
\AtBeginEnvironment{tabular}{\small}
\def\paperID{12495} 
\def\confName{CVPR}
\def\confYear{2025}

\newcommand{\cmark}{\ding{51}}%
\newcommand{\mybenchmark}{\textit{HalLoc}\xspace}
\newcommand{\mybenchmarks}{\textit{HalLoc-VQA}\xspace}
\newcommand{\mybenchmarkp}{\textit{HalLoc-Caption}\xspace}
\newcommand{\mybenchmarki}{\textit{HalLoc-Instruct}\xspace}
\newcommand{\mysize}{155K\xspace}

\newcommand{\mymodel}{\textit{HalLocalizer}\xspace}
\newcommand{\mypipeline}{\textit{HQA-Injection Pipeline}\xspace}

\title{\textit{HalLoc}: Token-level Localization of Hallucinations for Vision Language Models}

\author{
Eunkyu Park\thanks{Equal contribution.} \quad 
Minyeong Kim\footnotemark[1] \quad 
Gunhee Kim \thanks{Corresponding author.}\\ 
Seoul National University\\
{\tt\small eunkyu.park@vision.snu.ac.kr, \{kmy17518,gunhee\}@snu.ac.kr}}

\begin{document}
\maketitle
\begin{abstract}
Hallucinations pose a significant challenge to the reliability of large vision-language models, making their detection essential for ensuring accuracy in critical applications. Current detection methods often rely on computationally intensive models, leading to high latency and resource demands. Their definitive outcomes also fail to account for real-world scenarios where the line between hallucinated and truthful information is unclear. To address these issues, we propose HalLoc, a dataset designed for efficient, probabilistic hallucination detection. It features 150K token-level annotated samples, including hallucination types, across Visual Question Answering (VQA), instruction-following, and image captioning tasks. This dataset facilitates the development of models that detect hallucinations with graded confidence, enabling more informed user interactions. Additionally, we introduce a baseline model trained on HalLoc, offering low-overhead, concurrent hallucination detection during generation. The model can be seamlessly integrated into existing VLMs, improving reliability while preserving efficiency. The prospect of a robust plug-and-play hallucination detection module opens new avenues for enhancing the trustworthiness of vision-language models in real-world applications. The HalLoc dataset and code are publicly available at: \url{https://github.com/dbsltm/cvpr25_halloc}.
\end{abstract}
\section{Introduction}
\label{sec:intro}
Vision-Language Models (VLMs) \citep{ye2023mplugowl, liu2023visual, zhu2023minigpt4} harness the language generation capabilities of Large Language Models (LLMs) \citep{achiam2023gpt, touvron2023llama}, along with the visual understanding provided by vision foundation models \citep{dosovitskiy2021image, wang2023internimage}, to comprehend and describe visual content in natural language. Although VLMs have shown remarkable performance across various multimodal tasks, they are prone to producing descriptions that are inconsistent with the visual content, a phenomenon commonly referred to as \textit{hallucination} \citep{rohrbach2019object, liu2023visual, wang2023llmfree, li2023evaluating}.

\begin{figure}[tb]
  \centering
  \includegraphics[width=\linewidth]{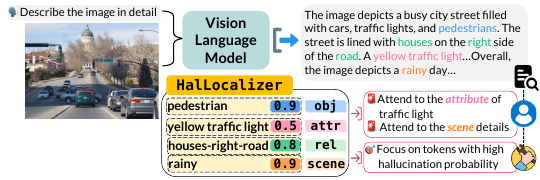}
  \caption{\textbf{Motivation behind HalLoc}: HalLoc aims to provide a practical, lightweight add-on for VLMs that simultaneously detects probabilistic hallucinations during response generation. By localizing tokens with high hallucination probabilities in real time, our model enhances use interaction by alerting at potentially unreliable content, prompting closer scrutiny or prioritizing external verification.}
  \label{fig:fig1_comp}
\end{figure}

Previous studies \citep{li2023evaluating, sun2023aligning, fu2024mme, wang2024amberllmfreemultidimensionalbenchmark, wu2024evaluatinganalyzingrelationshiphallucinations, qiu2024valorevalholisticcoveragefaithfulness, nie2024mmrelrelationunderstandingdataset} have explored the susceptibility of VLMs to various types of hallucinations, including those involving objects, visual attributes, spatial relations, and environmental details. The prevalence and diversity of these hallucinations pose significant challenges to real-world deployment, since they can lead to misinformation and undermine user trust \citep{biten2021let}.

Current approaches to hallucination detection often  leverages powerful off-the-shelf models like GPT-4 \cite{openai2024gpt4} to cross-verify outputs \citep{liu2024mitigatinghallucinationlargemultimodal, jing2024faithscorefinegrainedevaluationshallucinations, chen2024unifiedhallucinationdetectionmultimodal}. While effective, these methods introduce substantial computational overhead and latency, bringing challenges to their applications requiring real-time feedback or operating under stringent computational constraints. Moreover, existing methods typically yield definitive outputs as either \textit{hallucinated} or \textit{truthful}. This dichotomy fails to capture the nuanced spectrum of confidence levels needed for ambiguous cases where the distinction is unclear.

        


\begin{table*}[h!]
\centering
\caption{\textbf{Comparison of HalLocalizer} with existing hallucination detection methods.  Note that \textit{Discrete} output refers to definitive classes for detection and \textit{Probability} output stands for probability as confidence levels of hallucinations. }\label{tab:comparison1}
\begin{adjustbox}{max width=\textwidth}
\large
\begin{tabular}{l|cccccc}
        \toprule
        \multirow{2}{*}{Methods} & \multirow{2}{*}{Granularity} & \multirow{2}{*}{Type Classification} & \multirow{2}{*}{Detection} & \multicolumn{2}{c}{Output}& \multirow{2}{*}{VLM Tasks}\\ 
        \cmidrule[0.5pt](rl){5-6}
        &&&&Discrete&Probability& \\
        \hline
        \midrule
        \textbf{CHAIR} \citep{rohrbach2019object} &  Token & Object & Hand-crafted Rule & \cmark & & Img Captioning\\ 
        \textbf{GAVIE} \citep{liu2024mitigatinghallucinationlargemultimodal} &  Response & Object & GPT \cite{openai2024gpt4} as judge& \cmark& & Instruction Following \\
        \textbf{FAITHScore} \citep{jing2024faithscorefinegrainedevaluationshallucinations} & Atomic Fact & None & off-the-shelf & \cmark&& VQA, Img Captioning\\
        \textbf{UNIHD} \citep{chen2024unifiedhallucinationdetectionmultimodal} & Segment & Obj, Attr, Scene-text, Fact  &  off-the-shelf & \cmark&& VQA, Img Captioning\\ 
        \textbf{MHalDetect} \citep{gunjal2024detecting} & Sentence & None & Trained Model& &\cmark& Img Captioning \\ 
        \rowcolor{gray!20}
        \textbf{\mymodel (ours)} & Token & Obj, Attr, Rel, Sce & Trained Model&  &\cmark& \begin{tabular}{@{}c@{}}VQA, \\ Instruction Following,\\ Img Captioning \end{tabular}\\
        
        \bottomrule
    \end{tabular} 

\end{adjustbox}

\label{tab:halloc_evaluation}
\end{table*}

\begin{table}[tb]
\caption{\textbf{Comparison of HalLoc} with existing hallucination detection datasets.} \label{tab:comparison2}
\begin{adjustbox}{max width=\linewidth}\centering
    \resizebox{\textwidth}{!}{
    \large
   \begin{tabular}{l|cccc} \toprule 
Benchmarks & Size & Granularity & Type Annotation & VLM Tasks\\  \hline
\midrule
\textbf{HaELM}  \citep{wang2023evaluationanalysishallucinationlarge} & 56K & Response & None  & Img Captioning \\ 
\cmidrule(lr){2-5} 
\textbf{MHalDetect} \citep{gunjal2024detecting} & 16K & Segment & None & Img Captioning \\ 
\cmidrule(lr){2-5} 
\textbf{MHaluBench} \citep{chen2024unifiedhallucinationdetectionmultimodal}& 400 & Segment & 
\begin{tabular}{@{}c@{}} Obj, Attr,\\ Scene-text, Fact\end{tabular} & 
\begin{tabular}{@{}c@{}} VQA, \\ Img Captioning\end{tabular} \\
\cmidrule(lr){2-5} 
\rowcolor{gray!20}
\textbf{\mybenchmark (ours)} & \mysize & Token & 
\begin{tabular}{@{}c@{}}Obj, Attr \\ Rel, Sce\end{tabular}& 
\begin{tabular}{@{}c@{}}VQA, \\ Img Captioning, \\ Instruction Following \end{tabular} \\ \hline
\end{tabular}}
\end{adjustbox}
\end{table}

Therefore, there is a strong need for an efficient hallucination detection mechanism that offers probabilistic assessments. A plug-and-play hallucination detection module, as illustrated in Figure \ref{fig:fig1_comp}, can be seamlessly integrated into existing VLMs. Such a system enables more nuanced interactions with model outputs, benefiting users and downstream applications. For instance, users can be alerted to content requiring additional scrutiny without dismissing it outright, and systems can prioritize external verification for content with a higher likelihood of being hallucinated.

To address this need, we introduce HalLoc, a large-scale dataset comprising 155K samples spanning visual question answering (VQA), instruction following, and image captioning tasks. Each sample is meticulously annotated at the token level with specific types of hallucinations, facilitating comprehensive training and evaluation of hallucination detection models.

Leveraging HalLoc, we develop a simple yet effective baseline model that performs concurrent hallucination detection with minimal computational overhead. Our model can be integrated into existing Vision-Language Models (VLMs) in a plug-and-play fashion, enhancing their reliability without sacrificing efficiency.

Our contributions can be outlined as follows:
\begin{enumerate}
    \item Introduction of HalLoc dataset: We present HalLoc, a large-scale dataset comprising 155,000 samples annotated at the token level for specific types of hallucinations. Spanning tasks like visual question answering, instruction following, and image captioning, HalLoc facilitates comprehensive training and evaluation of hallucination detection models in VLMs.
    \item Development of an efficient hallucination detection Model: Leveraging HalLoc, we develop a simple yet effective baseline model that performs concurrent hallucination detection with minimal computational overhead. Unlike existing methods that provide binary outputs, our model offers probabilistic assessments, capturing nuanced confidence levels for ambiguous cases.
    \item Plug-and-Play integration with existing VLMs: Our hallucination detection model is designed to be seamlessly integrated into existing VLMs in a plug-and-play fashion. This enhances the reliability of VLMs without sacrificing efficiency, addressing the need for an efficient hallucination detection mechanism suitable for real-world applications with computational constraints.
\end{enumerate}

\section{Related Work}

\paragraph{Hallucination Detection Methods} Early work on hallucination detection focused on identifying non-existent objects in generated outputs. CHAIR \citep{rohrbach2019object} detects hallucinated objects by comparing mentioned objects with ground-truth data, using a simple rule-based approach limited to object-level hallucinations. Later approaches use powerful off-the-shelf models for detecting hallucinations. GAVIE~\citep{liu2024mitigatinghallucinationlargemultimodal} employs GPT-4~\citep{openai2024gpt4} to assign single accuracy scores to responses. FAITHScore~\citep{jing2024faithscorefinegrainedevaluationshallucinations} uses a chain of large models to break down responses into atomic facts and verify them via visual entailment. UNIHD~\citep{chen2024unifiedhallucinationdetectionmultimodal} identifies and classifies hallucinations at the semantic segment level by also leveraging a chain of off-the-shelf models including GPT-4~\cite{openai2024gpt4}.

HaELM~\citep{wang2023evaluationanalysishallucinationlarge} and MHalDetect~\citep{gunjal2024detecting} train models on specialized datasets to detect hallucinations at the sentence or response level. However, these approaches lack granularity as they detect hallucinations on long caption segments (inclusive of both hallucinated and non-hallucinated tokens) and do not provide hallucination type annotations, limiting their versatility. As shown in Table \ref{tab:comparison1}, unlike previous approaches, our \mymodel trained on \mybenchmark identifies object, attribute, relationship, and scene hallucinations at the token level across diverse tasks. HalLoc's confidence-based detection allows precise maneuvering over ambiguous hallucination tokens. Also, its lightweight design allows for a seamless integration with VLMs. 
\paragraph{Hallucination Detection Datasets} 
Previous works have crafted datasets to train automated hallucination detection models. HaELM~\citep{wang2023evaluationanalysishallucinationlarge} and MHalDetect~\citep{gunjal2024detecting} provide datasets with 56K and 16K samples, respectively, for image captioning with response-level or segment-level annotations. However, they do not verify over diverse tasks or include type-specific annotations, which are essential for a more comprehensive evaluation. MHaluBench~\citep{chen2024unifiedhallucinationdetectionmultimodal} provides a broader evaluation framework featuring 400 annotated samples that span image captioning and VQA, with segment-level annotations for objects, attributes, scene-text, and fact hallucinations. However, its relatively small size makes it more suited for evaluation than large-scale model training. 
As illustrated in Table~\ref{tab:comparison2}, HalLoc distinguishes itself as the only dataset offering token-level granularity with annotations for objects, attributes, relationships, and scene types, supporting the broadest array of vision-language tasks. Despite the complexity of granular annotations, it remains the largest dataset with 155K samples, providing a strong foundation for training versatile automated hallucination detection models like HalLocalizer.

  
  

\begin{figure*}[t!]
  \centering
  \includegraphics[width=\linewidth]{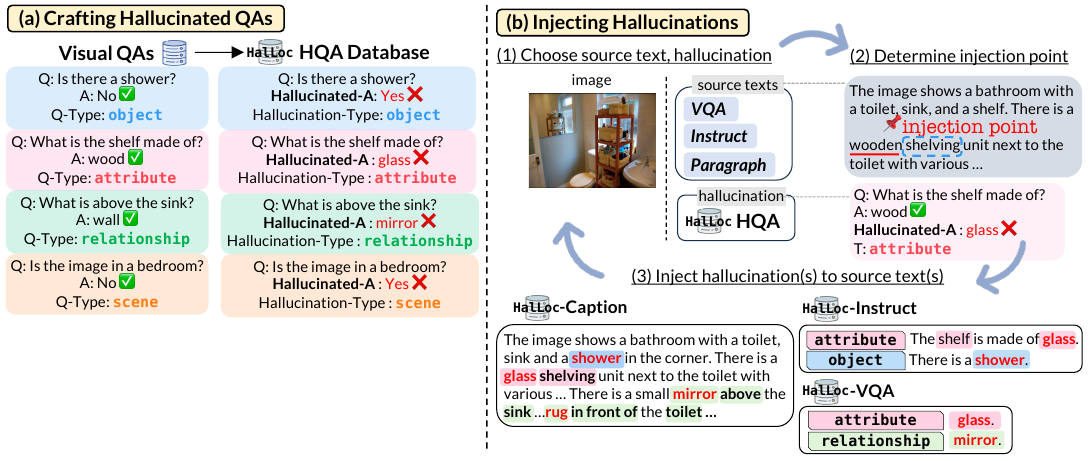}
  \caption{Overview of our automated hallucination annotation pipeline with HQA Injection. The pipeline consists of two stages: hallucination generation and injection. First, we (a) \textbf{craft} (question, hallucinated answer) pairs for each hallucination type, using strategies reflecting concept association biases statistical biases in VLMs (details are shown \S~\ref{sec:generating_hall}). Then (b) GPT-4 \citep{openai2024gpt4} \textbf{injects} hallucinated QA pairs from the HQA Database into source texts that refer to the same image. Source texts span VQA, instructions, and paragraphs and vary for the three task types of \mybenchmark via text-based QA. Details for each source tasks description for each task is discussed in \S~\ref{sec:composition_tasks}.}
  \label{fig:pipeline}
\end{figure*}

\section{Hallucination Types and Patterns}
\subsection{Hallucination Types}
\label{sec:hallucination_type}
Previous studies \citep{sun2023aligning, fu2024mme, wang2024amberllmfreemultidimensionalbenchmark, wu2024evaluatinganalyzingrelationshiphallucinations} highlight that VLMs exhibit diverse forms of hallucinations, underscoring the need for comprehensive coverage. In \mybenchmark, we address four types of hallucinations caused by image-text misalignment: object, attribute, relation, and scene. To enable token-level detection, we provide a detailed taxonomy.

\paragraph{Object Hallucination {\normalfont (\texttt{<obj>})}}
A non-existent object. Only the token referring to the misidentified object constitutes an object hallucination. Attributes or relationships associated with the nonexistent object are excluded.

\paragraph{Attribute Hallucination {\normalfont (\texttt{<attr><obj>})}}
An incorrect property that is associated with a single object. Properties may be visual features, spatial positions, or actions involving a single object (e.g., \textit{white shelf}, \textit{door to the left of the image}, \textit{sleeping woman}). Both the object token and the property token constitute an attribute hallucination.

\paragraph{Relationship Hallucination {\normalfont (\texttt{<obj1><rel><obj2>})}}
An unfaithful interaction involving two objects. Interactions include relative spatial positions or actions between two objects (e.g., \textit{mirror above the sink}, \textit{man holding an apple}). The two object tokens and the interaction token constitute a relationship hallucination.

\paragraph{Scene Hallucination {\normalfont (\texttt{<sce>})}}
An incorrect scene description of the image. Scenes include weather, place, or location (e.g., \textit{sunny}, \textit{office}, \textit{outdoors}). Only the scene token constitutes a scene hallucination. 

\subsection{Hallucination Patterns}
\label{sec:hallucination_pattern}
Accurate representation of hallucination patterns is vital to the quality of the dataset. Below, we outline common patterns guiding \mybenchmark's generation pipeline.

\paragraph{Concept Association Bias}
Concept Association Bias (CAB)~\citep{yamada2024lemons} occurs when vision-language models treat an input as a bag of concepts and attempt to infer missing concepts cross-modally. For example, if an image contains both a lemon and an eggplant, the model might incorrectly predict the lemon as purple, influenced by the strong semantic connection between \textit{purple} and \textit{eggplant}, rather than accurately relying on the visual cues. 

\begin{table}[h]
\caption{VQA results for probing statistical biases in hallucinations. We pose binary questions modeling VLMs to elicit hallucinations stemming from \textbf{Lang}uage or \textbf{Image} priors. The questions ask about attribute, relationship, scene-type hallucinations likely to co-occur due to these priors. \textbf{A}(ccuracy), \textbf{P}(recision), \textbf{R}(ecall), and \textbf{F1}(-score) metrics are reported. Yes \% denotes the proportion of questions answered with \enquote{Yes}.} \label{tab:bias_analysis_yesno}
\begin{adjustbox}{max width=\linewidth}\centering
    \resizebox{\textwidth}{!}{
    \large
   \begin{tabular}{@{}cl|ccccc@{}}
    \toprule
     Hallucination Type & Prior & \textbf{A} & \textbf{P} & \textbf{R} & \textbf{F1} & Yes\% \\ \midrule[0.7pt]
     \multirow{2}{*}{Attribute} & Lang & 0.47 &0.45 & 0.29 & 0.36 & 0.32 \\
     & Image & 0.52 &0.53 & 0.33 &0.41 &0.31\\ \cline{1-7}
    \multirow{2}{*}{Relationship}& Lang & 0.71 &0.69&0.75 &0.72 & 0.53 \\
    & Image & 0.64 &0.61&0.74&0.67&0.60 \\ \cline{1-7}
     \multirow{2}{*}{Scene}& Lang & 0.35 &0.41&0.66&0.50 &0.81\\
     & Image & 0.33&0.39&0.63&0.48&0.80 \\ \bottomrule
  \end{tabular}}
\end{adjustbox}

\end{table}

    
\paragraph{Statistical Bias}
Hallucinations in VLMs have also been attributed to an overreliance on statistical bias~\citep{agarwal2020causalvqarevealingreducing} in unimodal priors~\citep{gupta2022swapmixdiagnosingregularizingoverreliance, han2022visualperturbationawarecollaborativelearning}. 

\noindent
(1) \textit{Language Prior}: The model generates a hallucinated response due to textual context alone, represented as
$P(A_{\text{hal}} \mid Q) > P(A \mid Q)$.

\noindent
(2) \textit{Image Prior}: The model generates a hallucinated response due to the visual context, represented as
$P(A_{\text{hal}} \mid I) > P(A \mid I)$. 

\noindent
(3) \textit{Language-Image Prior}: The model generates a hallucinated response due to both the textual and visual context, represented as
$P(A_{\text{hal}} \mid I,Q) > P(A \mid I,Q)$. 

To demonstrate that VLMs exhibit these statistical biases, we pose binary (yes/no) questions targeting frequent attribute, relationship, and scene combinations. The binary questions are of the form: \enquote{Is \texttt{<obj> <wrong-attribute>}?} (e.g., \textit{Is the lemon purple?}), \enquote{Is \texttt{<wrong-subj> <predicate> <obj>}?}, \enquote{Is \texttt{<subj> <wrong-predicate> <obj>}?} The question set includes an equal number of questions with \enquote{Yes} and \enquote{No} answers. VQA results in Table~\ref{tab:bias_analysis_yesno} show low performance in discerning highly co-occurring attributes and scenes. For scene hallucinations, the high \enquote{Yes} rate indicates that the model is overconfident, resulting in lower accuracy on questions where the correct answer is \enquote{No}.

\section{HalLoc}
We present \mybenchmark, the first large-scale benchmark (155K samples) with token-level annotations of hallucination types spanning three core vision-language tasks: Visual Question Answering (VQA), Instruction-following, and Image Captioning. 

\subsection{Generation Pipeline}
\label{sec:generating_hall}
The comprehensive scale and fine-grained annotations of \mybenchmark are enabled by our \mypipeline, which allows for controllable and scalable hallucination annotation. The \mypipeline is a two-stage process: crafting hallucinated answers and injecting hallucinations. An overview of our \mypipeline is illustrated in Figure~\ref{fig:pipeline}. 

\paragraph{Crafting Hallucinated Answers}
We first construct a large Hallucinated Question-Answer (HQA) database, which contains questions, their corresponding hallucinated answers, and annotations of the hallucination types based on the taxonomy defined in Section~\ref{sec:hallucination_type}.
The questions are sourced from the GQA dataset~\citep{hudson2019gqa}, a Visual Question Answering benchmark featuring images from the Visual Genome dataset~\citep{krishna2016visual}. We select questions pertaining to the existence of objects, object attributes, relationships between objects, and environmental surroundings. Each question is labeled with its corresponding type (object, attribute, relationship, or scene).

Next, we employ two strategies to craft hallucinated answers for each question, aiming to mimic real-world hallucination patterns.

\noindent
(1) We use GPT-4~\citep{openai2024gpt4} to select from hallucinated answer candidates originating from hallucination patterns observed in Vision-Language Models (VLMs), as defined in Section~\ref{sec:hallucination_pattern}. For Concept Association Bias, we collect hallucinated answer candidates by sampling traits of the same category associated with different objects in the image. For example, in Figure~\ref{fig:pipeline}~(a), the attribute hallucination \textit{glass shelf} is created by borrowing the attribute \textit{glass} from the \textit{glass window} present in the image. For Statistical Bias, we gather data on the co-occurrence of objects, attributes, relationships, and scenes associated with a particular object. The candidates are subsequently categorized into specific types: language-prior, image-prior, or language-image-prior. 

\noindent
(2) We additionally use GQA~\citep{hudson2019gqa} decoys or responses from VLMs. To capture hallucinations beyond the predefined types, we directly employ decoy answers or model-generated responses as hallucinated answers. Some GQA questions naturally have decoys (e.g., question: \enquote{Is the color of the shoe blue?} answer: \enquote{No} where color \textit{blue} reflects concept association bias or co-occurrence biases), which we use as hallucinated answers. Furthermore, we conduct VQA on a subset of the selected questions using two popular VLMs: LLaVA 1.5 (7B)~\citep{liu2023visual} and InstructBLIP (7B)~\citep{dai2023instructblip}. After filtering out correct answers via exact match, a response is accepted as a hallucinated answer only if three strong VLMs unanimously agree that the response is not visually entailed. The three LVLMs used for verification are InternVL 2.0~\citep{chen2024internvl} (26B), LLaVA 1.5 (13B), and InstructBLIP (13B).

Finally, following our defined taxonomy, we annotate the complete hallucination traits in each question-hallucinated answer pair. The type of the question naturally determines the hallucination type. The scene graphs provided in the GQA dataset allow for algorithmic annotations, which are subsequently verified by GPT-4 to eliminate any erroneous annotations.

\paragraph{Injecting Hallucinations}
With the large-scale HQA Database containing diverse types of hallucinated answers, we proceed to craft hallucinated responses for various tasks by systematically injecting these hallucinations.

First, we select source texts and questions from the HQA Database. The source texts are drawn from existing datasets designed for a variety of vision-language tasks and include prompts describing the task, accompanied by ground-truth responses free of hallucinations. We sample questions from the HQA database such that each source text and question shares the same reference image.

Next, we identify the injection point by querying the ground-truth response with the chosen question using GPT-4.  If an answer is detected within the text, its position is designated as the injection point. If no answer is found, GPT-4 suggests a natural injection point within the text.

Finally, we inject the hallucinated answer into the text. If an answer was previously identified, it is replaced with the hallucinated answer. If no answer is found, GPT-4 generates a phrase or sentence containing the hallucinated answer and inserts it at the injection point. We annotate the complete component of the underlying hallucination trait. This process is challenging, even for a robust model such as GPT-4; therefore, we incorporate an additional verification step. For each question, the process is repeated up to three times until GPT-4 verifies that the injection is acceptable.

\begin{figure}[!ht] \centering
    
    \includegraphics[width=\linewidth]{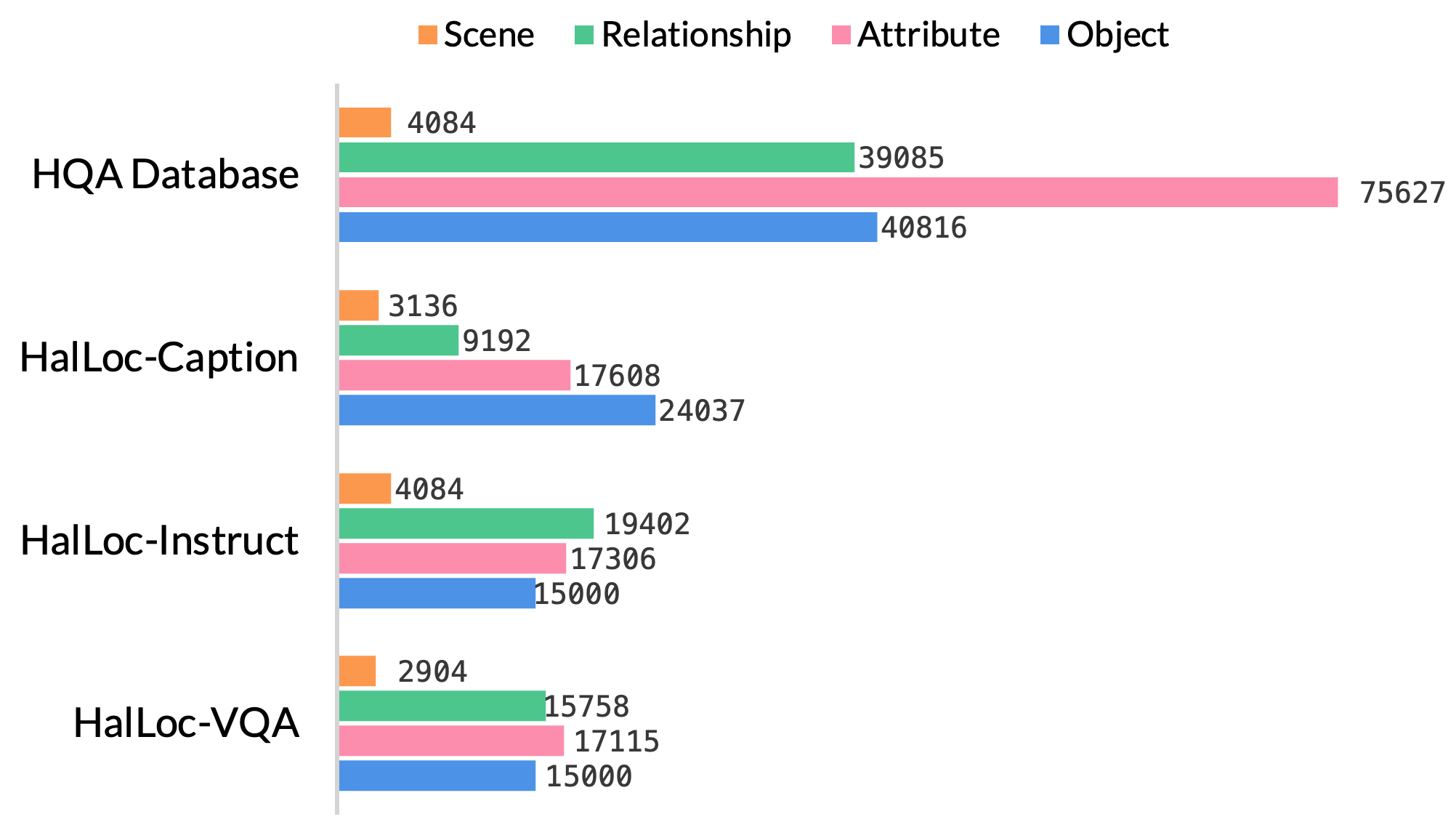} 
    \caption{Number of data points per each hallucination type in the HQA Database, \mybenchmarks, \mybenchmarki, \mybenchmarkp.}
    \label{fig:type_dist}
\end{figure}

\subsection{HalLoc Composition}
\label{sec:composition_tasks}
\paragraph{HQA Database}

The HQA Database comprises 159,612 questions. Regarding hallucination types, there are 40,816 object-hallucinated answers, 75,627 attribute-hallucinated answers, 39,085 relationship-hallucinated answers, and 4,084 scene-hallucinated answers. The relative scarcity of scene hallucinations is attributable to the limited number of images with clearly discernible environmental settings, such as weather or location. The distribution of hallucination types across each subset of \mybenchmark\ is illustrated in Figure~\ref{fig:type_dist}. In terms of hallucination patterns, there are 15,524 Concept Association Bias hallucinated answers, 42,671 language-image prior hallucinated answers, 16,546 language prior hallucinated answers, and 64,129 hallucinated answers derived directly from GQA decoys or incorrect VLM responses.

\paragraph{HalLoc}

HalLoc comprises 155,953 samples, divided into 124,528 training samples, 16,271 validation samples, and 15,154 test samples. Each sample contains an average of 17.66 words, with an average of 1.81 hallucinated words (about 10\%). For comparative purposes, HalLoc also includes 13,713 non-hallucinatory samples. HalLoc is partitioned into three subsets: \mybenchmarks, \mybenchmarki, and \mybenchmarkp. 

\paragraph{\mybenchmarks} This subset contains 55,854 samples. The source text is derived from ground-truth answers in the GQA dataset. As it models the Visual Question Answering (VQA) task, the average number of words per sample is notably small—1.05 words, with an average of 0.96 hallucinated words (91\%). It includes 15,000 object hallucinations, 17,115 attribute hallucinations, 15,758 relationship hallucinations, and 2,904 scene hallucinations.

\paragraph{\mybenchmarki} This subset contains 60,862 samples. The source text is drawn from sentence claims in the GQA dataset. Additionally, GPT-4 generates prompts tailored to each sentence to model diverse instructions. The average number of words per sample is 7.21, with an average of two hallucinated words (28\%). It includes 15,000 object hallucinations, 17,306 attribute hallucinations, 19,402 relationship hallucinations, and 4,084 scene hallucinations.

\paragraph{\mybenchmarkp}

This subset contains 39,237 samples. The source text originates from the Stanford Image Paragraphs~\citep{krause2016paragraphs} and Localized Narratives~\citep{ponttuset2020connecting} datasets. It is the longest subset, with an average of 57.53 words per sample and 2.72 hallucinated words (5\%). It is also the most diverse in terms of the number of injected hallucinations per source text, ranging from 0 to 6. It includes 24,037 object hallucinations, 17,608 attribute hallucinations, 9,192 relationship hallucinations, and 3,136 scene hallucinations. The relative imbalance between object, attribute, and relationship hallucinations compared to other subsets arises from higher failure rates during hallucination injection for relationship types. 


\section{\mymodel}
\begin{figure}
    \centering
    
    \includegraphics[width=\linewidth]{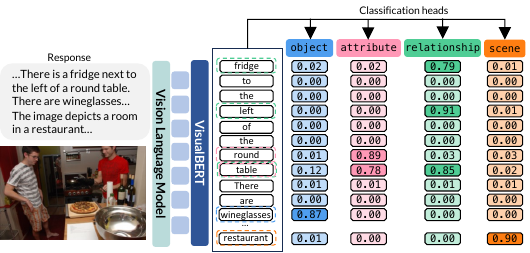}
    \caption{\mymodel employs a bidirectional VisualBERT encoder and 4 linear classification heads for token-level localization of each four hallucination types.}
    \label{fig:model_arc}
\end{figure}
\begin{table*}[htb]
\centering
\caption{Token-level hallucination detection results on HalLoc. We compare HalLocalizer with existing methods across three tasks: VQA, Instruction Following, and Image Captioning. For each hallucination type, we report Precision (P), Recall (R), and F1-score (F1). CHAIR and Always-1 are baseline methods.}
\label{tab: halloc_performance}
\begin{adjustbox}{max width=\textwidth}
\begin{tabular}{@{}c|ll|cccccccccccc@{}}
\toprule
\multirow{2}{*}{Subset} & \multirow{2}{*}{Method} & \multirow{2}{*}{LVLM Embedding} & \multicolumn{3}{c}{Object} & \multicolumn{3}{c}{Attribute} & \multicolumn{3}{c}{Relationship} & \multicolumn{3}{c}{Scene} \\
\cmidrule(lr){4-6} \cmidrule(lr){7-9} \cmidrule(lr){10-12} \cmidrule(lr){13-15}
 & & & P & R & F1 & P & R & F1 & P & R & F1 & P & R & F1 \\
\midrule
\multirow{7}{*}{VQA} 
 & CHAIR & - & 0.27 & 0.15 & 0.19 & - & - & - & - & - & - & - & - & -  \\
 & Always 1 & - & 0.28 & 1.00 & 0.44 & 0.28 & 1.00 & 0.44 & 0.28 & 1.00 & 0.44 & 0.06 & 1.00 & 0.12 \\
 \cmidrule(lr){2-15}
 & \multirow{5}{*}{HalLocalizer} & MiniGPT4 & 0.54 & 0.87 & 0.67 & 0.83 & 0.91 & 0.87 & 0.53 & 0.82 & 0.64 & 0.86 & 0.75 & 0.80 \\
 &  & InstructBLIP & 0.54 & 0.74 & 0.62 & 0.63 & 0.78 & 0.69 & 0.42 & 0.83 & 0.56 & 0.9 & 0.80 & 0.84  \\
 &  & LLaVA & 0.57 & 0.87 & 0.69 & 0.87 & 0.95 & 0.90 & 0.54 & 0.83 & 0.66 & 0.94 & 0.85 & 0.89 \\
 &  & InternVL & 0.61 & 0.87 & \cellcolor{gray!20}\textbf{0.71} & 0.91 & 0.96 & \cellcolor{gray!20}\textbf{0.94} & 0.60 & 0.86 & \cellcolor{gray!20}\textbf{0.71} & 0.94 & 0.93 & \cellcolor{gray!20}\textbf{0.93} \\
 &  & w/o Embedding & 0.59 & 0.82 & 0.69 & 0.71 & 0.83 & 0.77 & 0.48 & 0.77 & 0.60 & 0.78 & 0.84 & 0.81 \\
\midrule
\multirow{7}{*}{Instruct}
 & CHAIR & -& 0.15 & 0.12 & 0.13 & - & - & - & - & - & - & - & - & -  \\
 & Always 1 & - & 0.04 & 1.00 & 0.07 & 0.08 & 1.00 & 0.15 & 0.14 & 1.00 & 0.26 & 0.01 & 1.00 & 0.02 \\
 \cmidrule(lr){2-15}
 & \multirow{5}{*}{HalLocalizer} & MiniGPT4 & 0.89 & 0.68 & 0.77 & 0.90 & 0.95 & 0.92 & 0.68 & 0.81 & 0.74 & 0.91 & 0.96 & 0.93 \\
 &  & InstructBLIP & 0.88 & 0.70 & 0.78 & 0.91 & 0.95 & 0.93 & 0.73 & 0.90 & 0.81 & 0.91 & 0.95 & 0.93 \\
 &  & LLaVA & 0.87 & 0.72 & 0.79 & 0.93 & 0.95 & 0.94 & 0.75 & 0.89 & 0.81 & 0.96 & 0.93 & \cellcolor{gray!20} \textbf{0.94} \\
 &  & InternVL & 0.89 & 0.71 & 0.79 & 0.94 & 0.96 & 0.95 & 0.75 & 0.95 & \cellcolor{gray!20} \textbf{0.84} & 0.94 & 0.93 & \cellcolor{gray!20}\textbf{0.94}  \\
 &  &  w/o Embedding & 0.94 & 0.73 & \cellcolor{gray!20}\textbf{0.82} & 0.94 & 0.99 & \cellcolor{gray!20}\textbf{0.97} & 0.75 & 0.92 & \cellcolor{gray!20}\textbf{0.83} & 0.93 & 0.94 & \cellcolor{gray!20}\textbf{0.94}  \\
\midrule
\multirow{7}{*}{Caption}
 & CHAIR & - & 0.03 & 0.08 & 0.04 & - & - & - & - & - & - & - & - & -  \\
 & Always 1 & - & 0.01 & 1.00 & 0.02 & 0.02 & 1.00 & 0.02 & 0.01 & 1.00 & 0.01 & 0.002 & 1.00 & 0.005 \\
 \cmidrule(lr){2-15}
 & \multirow{5}{*}{HalLocalizer} & MiniGPT4 & 0.26 & 0.27 & 0.27 & 0.15 & 0.19 & 0.17 & 0.33 & 0.29 & 0.31 & 0.58 & 0.35 & 0.43 \\
 &  & InstructBLIP & 0.37 & 0.38 & 0.37 & 0.26 & 0.29 & 0.28 & 0.52 & 0.44 & 0.47 & 0.62 & 0.51 & 0.56 \\
 &  & LLaVA & 0.47 & 0.40 & 0.43 & 0.31 & 0.27 & 0.29 & 0.56 & 0.39 & 0.46 & 0.73 & 0.48 & 0.58 \\
 &  & InternVL & 0.56 & 0.60 & 0.58 & 0.42 & 0.34 & 0.37 & 0.52 & 0.41 & 0.46 &0.26 & 0.25 & 0.25 \\
  &  &  w/o Embedding & 0.66 & 0.70 & \cellcolor{gray!20}\textbf{0.68} & 0.67 & 0.60 & \cellcolor{gray!20}\textbf{0.64} & 0.71 & 0.71 & \cellcolor{gray!20}\textbf{0.71} &0.85 & 0.69 & \cellcolor{gray!20}\textbf{0.76} \\
\bottomrule
\end{tabular}
\end{adjustbox}
\end{table*}

The goal of \mybenchmark is to support the development of plug-and-play hallucination detection modules that can be seamlessly integrated into existing Vision-Language Models (VLMs). As an initial exploration in this direction, we introduce \textbf{\mymodel}, a family of lightweight token-level hallucination detection models trained on \mybenchmark.

\subsection{Task}
\label{sec:localization_task}
The task of \mymodel is to identify the token-level location and type of hallucinations in arbitrary model responses for a given image. Each token is evaluated to determine whether it constitutes a hallucination for each type. Formally, let \( H(t) \) represent the label for token \( t \), where \( H(t) \in \{0, 1\} \) indicates whether token \( t \) is a hallucination (1) or not (0). 
For each type \( h \), a localization model predicts a function \( H_h(t) \), which determines whether a given token \( t \) is classified as a hallucination of type \( h \), $h \in \{\text{object}, \text{attribute}, \text{relationship}, \text{scene}\}$.

\subsection{Architecture}

\mymodel\ features two primary architectures designed to support immediate detection of token-level hallucinations, including confidence levels, during response generation.

The first architecture leverages the vision-language model embeddings—specifically, the last hidden state of the vision-language model during response generation—as input to the detection module.
The second architecture directly utilizes the text response generated by the vision-language model as input to the detection module.
In both cases, the detection module consists of a VisualBERT ~\citep{li2019visualbertsimpleperformantbaseline} encoder followed by four linear classification heads, one for each type of hallucination. 

An illustration of our proposed architecture is provided in Figure~\ref{fig:model_arc}. When \mymodel\ is integrated into existing vision-language models, it enables them to generate responses alongside token-level hallucination probabilities for each hallucination type.

\subsection{Training Details}
In both architectures, only the detection modules are trainable—the parameters of the VisualBERT encoder and the four linear classification heads. For models that utilize LVLM embeddings, we cache these embeddings prior to training to improve efficiency. We set the sequence length to 512 in all experiments and fixed the batch size at 64. We train \mymodel\ using the AdamW optimizer~\citep{loshchilov2019decoupledweightdecayregularization} with $\beta=(0.9, 0.999)$, $\epsilon=1.0 \times 10^{-8}$, and a weight decay of $1.0 \times 10^{-2}$. Training is conducted for 25 epochs on four A6000 GPUs, resulting in an average training time of 10 hours. We employ a learning rate of $1.0 \times 10^{-6}$, with a cosine annealing scheduler applied across all trials.


\section{Experiments}
We evaluate the performance of \mymodel on the \mybenchmark dataset on each of the three subsets: VQA, Instruct, and Caption. The methods compared\footnote{Due to the inability of other detection methods to perform token-level hallucination detection, we cannot directly compare the performance.} include a naive baseline classifier that always predicts 1 (\textit{Always 1}), CHAIR~\citep{rohrbach2019object} (\textit{CHAIR})\footnote{CHAIR is limited to detecting object hallucinations only.}, and our proposed HalLocalizer (\textit{HalLocalizer}) with various Large Vision-Language Model (LVLM) embeddings—MiniGPT4~\citep{zhu2023minigpt4}, InstructBLIP~\citep{dai2023instructblip}, LLaVA~\citep{liu2023visual}, InternVL~\citep{chen2024internvl}—as well as without any LVLM embedding. The evaluation metrics\footnote{Each classification head's threshold is individually set to its optimal value based on the validation set for each subset.} are Precision (\textit{P}), Recall (\textit{R}), and F1-score (\textit{F1}) computed across four categories: Object, Attribute, Relationship, and Scene (see Table~\ref{tab: halloc_performance}).

\subsection{VQA}
HalLocalizer demonstrates superior performance, particularly when combined with LVLM embeddings. HalLocalizer with InternVL embedding attains the highest F1-scores across all categories: 0.71 for Object, 0.94 for Attribute, 0.71 for Relationship, and 0.93 for Scene.
\subsection{Instruct}
Remarkably, HalLocalizer without any LVLM embeddings achieves the highest F1-scores for Object (0.82) and Attribute (0.97), and ties for the best F1-score in Scene (0.94). For the Relationship category, both HalLocalizer with InternVL and without embedding perform competitively, with F1-scores of 0.84 and 0.83, respectively. 
\subsection{Caption}
HalLocalizer without any LVLM embedding outperforms all other configurations, achieving F1-scores of 0.68 for Object, 0.64 for Attribute, 0.71 for Relationship, and 0.76 for Scene. 

\section{Analysis}
\subsection{Results of Human Evaluation}
For human evaluation, we employ six AMT workers to assess hallucinated contents in HalLoc. We randomly sample 100 data points from each of the sub-benchmarks. For each benchmark type, workers are asked to rate each sample on two binary (yes/no) questions: 

\begin{table}[h!]\centering
\caption{Average accuracies across \mybenchmark. Workers assigned to each benchmark show unanimous agreement on their ratings.}
    \label{tab:human_eval}
    \begin{adjustbox}{max width=\linewidth}
    \resizebox{0.32\textwidth}{!}{
    \begin{tabular}{@{}l|cc@{}}
    \toprule
         & Generation& Injection  \\ \midrule[0.7pt]
         Accuracy & 0.91 & 0.98 \\ 
        Worker Agreement & 100 \% & 100 \% \\ \bottomrule
    \end{tabular}}
   \end{adjustbox}
   
\end{table}

\begin{enumerate}
    \item \textbf{Generation Accuracy}: Are the hallucinated tokens \textit{truly} hallucinated? (i.e., Do the hallucinated tokens convey false information in the context of the image?)
    \item \textbf{Injection Accuracy}: Are the hallucinated segments adequately injected into the source text in terms of context?
\end{enumerate}
We report the average accuracies across HalLoc in Table~\ref{tab:human_eval}. We ensure that AMT workers receive fair wages of approximately \$18 per hour.

\subsection{Limitations of Using Token Log Probabilities}
\begin{table}[tb]
\caption{Comparison of HalLocalizer and token-level log probabilitiy baselines across HalLoc. Total stands for binary hallucination labels, encompassing all hallucination types.}\label{tab:log_probability} \centering
\begin{adjustbox}{max width=\linewidth}
    \large
    \begin{tabular}{@{}cll|cccc@{}} \toprule
        \multirow{2}{*}{HalLoc Subset} & \multirow{2}{*}{Method} & \multirow{2}{*}{Model Arch} & \multicolumn{3}{c}{Total}  \\
        \cmidrule(lr){4-6}
        &&&P&R&F1 \\ \midrule
        \multirow{2}{*}{VQA} & HalLocalizer & \begin{tabular}{@{}c@{}}InternVL + \\ VisualBERT \end{tabular} &0.91 &1&0.95 \\
        \cmidrule(lr){2-7}
        & LogProb & LLaVA &0.9 &1&0.95 \\ \midrule
        \multirow{2}{*}{Instruct} & HalLocalizer & \begin{tabular}{@{}c@{}}VisualBERT \end{tabular} &0.86 &0.97&0.91& \\  \cmidrule(lr){2-7}
        & LogProb & LLaVA &0.32 &0.94 &0.47 \\ \midrule

        \multirow{2}{*}{Caption} & HalLocalizer & \begin{tabular}{@{}c@{}}VisualBERT \end{tabular} &0.72 &0.71&0.71& \\
        \cmidrule(lr){2-7}
        & LogProb & LLaVA &0.12 &0.30 &0.17 \\ \bottomrule
\end{tabular}
\end{adjustbox}
\end{table}

Log probabilities from VLMs can offer a simple alternative to lightweight hallucination detection modules. However, they come with notable limitations: First, log probabilities cannot distinguish between different types of hallucinations. Second, we empirically tested the effectiveness of token-level log probabilities in identifying hallucinated tokens using the HalLoc dataset. The results are summarized in Table~\ref{tab:log_probability}. Although a high log probability may numerically suggest confidence, without proper calibration, this confidence is not necessarily justified. Calibrated uncertainty from estimates from HalLocalizer adjusts these probabilities to better align with real-world correctness.

In the VQA subset, both methods perform comparably well, each achieving a high F1-score of 0.95, indicating strong capabilities in detecting hallucinations within question-answering contexts. However, in the Instruct subset, HalLocalizer significantly outperforms the LogProb method, achieving F1-scores of 0.91 and 0.47, respectively. This disparity is primarily attributed to the LogProb method's low precision of 0.32, suggesting that it produces a substantial number of false positives despite the high recall of 0.94.

In the Caption subset, the performance gap widens further: \mymodel achieves an F1-score of 0.71, whereas the LogProb method yields a considerably lower F1-score of 0.17. These results demonstrate that \mymodel consistently delivers more reliable hallucination detection across diverse tasks, particularly excelling in scenarios where the LogProb method struggles due to its inability to handle longer, more complex responses accurately. 

\subsection{Ablation Study on LVLM Embeddings}
\begin{table}[tb]
\centering
\caption{Ablation study on LVLM Embeddings}
\label{tab:7_1_LVLMembeddings}
\begin{adjustbox}{max width=\linewidth}
    \large
\begin{tabular}{@{}cl|cccc@{}}
\toprule
\multirow{2}{*}{HalLoc Subset} & \multirow{2}{*}{Method} & \multicolumn{1}{c}{Obj}& \multicolumn{1}{c}{Attr}& \multicolumn{1}{c}{Rel} & \multicolumn{1}{c}{Scen} \\ 
\cmidrule(lr){3-3} \cmidrule(lr){4-4} \cmidrule(lr){5-5} \cmidrule(lr){6-6}
& & F1 & F1 & F1 & F1 \\ \midrule
\multirow{3}{*}{VQA} & VisualBERT & 0.69 & 0.77 & 0.60 & 0.81 \\
 & LVLM & 0.72 & 0.95 & 0.70 & 0.93 \\
 & LVLM + VisualBERT & 0.71 & 0.94 & 0.71 & 0.93\\ \midrule
 \multirow{3}{*}{Instruct} & VisualBERT & 0.82 & 0.97 & 0.83 & 0.94 \\
 & LVLM & 0.68 & 0.95 & 0.84 & 0.95 \\
 & LVLM + VisualBERT  & 0.79 & 0.95 & 0.84 & 0.94\\ \midrule
 \multirow{3}{*}{Caption} & VisualBERT & 0.68 & 0.64 & 0.71 & 0.71 \\
 & LVLM & 0.58 & 0.41 & 0.54 & 0.65 \\
 & LVLM + VisualBERT  & 0.58 & 0.37 & 0.46 & 0.25\\
\bottomrule
\end{tabular}
\end{adjustbox}
\end{table}

To thoroughly examine the impact of using LVLM embeddings compared to textual inputs, we investigate three variants of HalLocalizer. We employ InternVL as the LVLM in these experiments, given its robust performance among models using LVLM embeddings. The first variant, \textit{VisualBERT}, processes the text responses generated by the vision-language model as input to the detection module and utilizes only its own vision encoder for visual information. The second variant, \textit{LVLM}, is specifically crafted for this study and replaces VisualBERT with a BERT encoder, relying exclusively on LVLM embeddings for visual input. The third variant, \textit{LVLM+VisualBERT}, feeds the embeddings generated by the vision-language model into the detection module, integrating both the LVLM embeddings and the VisualBERT vision encoder.

As shown in Table~\ref{tab:7_1_LVLMembeddings}, in the VQA subset, both \textit{LVLM} and \textit{LVLM+VisualBERT} outperform \textit{VisualBERT}, particularly in recognizing attributes and scenes, achieving F1-scores of up to 0.95 and 0.93, respectively. In contrast, in the Instruct and Caption subsets, \textit{VisualBERT} surpasses the other models, suggesting that LVLM embeddings may not provide additional benefits in hallucination detection for more extended responses. This observation may be related to a decrease in calibration as the response length increases, causing LVLM embeddings to convey uncertain information less reliably. However, definitive conclusions cannot be drawn at this stage and we leave this topic for future research. 

\section{Conclusion}
In this paper, we introduced HalLoc, a large-scale dataset comprising 155K samples across visual question answering, instruction-following, and image captioning tasks. Each sample is annotated at the token level with specific types of hallucination, allowing fine-grained training and evaluation. Using HalLoc, we developed baseline models capable of concurrent hallucination detection with minimal overhead. Unlike existing binary approaches, HalLocalizer provides probabilistic assessments, capturing confidence levels for ambiguous cases. This lightweight module can be seamlessly integrated into existing Vision-Language Models (VLMs), enhancing their reliability without compromising efficiency.

Our work lays the foundation for more advanced and lightweight hallucination detection modules, which also opens new directions for  future research on sophisticated hallucination mitigation strategies~\citep{zhou2024analyzing, kim2024exploitingsemanticreconstructionmitigate}. This scalable approach strikes a balance between accuracy and efficiency, advancing research in the practical deployment of trustworthy VLMs.

\section{Acknowledgments}
This work was supported by the SNU-Global Excellence Research Center establishment project, the National Research Foundation of Korea (NRF) grant funded by the Korea government (MSIT) (No.~2023R1A2C2005573), the Institute of Information \& Communications Technology Planning \& Evaluation (IITP) grant funded by the Korea government (MSIT) (No.~RS-2019-II191082, SW StarLab), the Institute of Information \& communications Technology Planning \& Evaluation (IITP) grant funded by the Korea government (MSIT) (No.~RS-2022-II220156, Fundamental research on continual meta-learning for quality enhancement of casual videos and their 3D metaverse transformation), and the Institute of Information \& communications Technology Planning \& Evaluation (IITP) grant funded by the Korea government (MSIT) (No.~RS-2021-II211343, Artificial Intelligence Graduate School Program (Seoul National University)). Gunhee Kim is the corresponding author.





\appendix
\section{Motivation for a Probabilistic Hallucination Detection Module}
By developing a model that outputs calibrated token-level probabilities of uncertainty, we gain a fine-grained understanding of the model's confidence in each word it produces. This calibration ensures that the probability estimates accurately reflect the true likelihood of correctness. When the model predicts a token with high uncertainty, it may indicate a higher risk of hallucination at that point in the text.

Even though we have token-level log probabilities, they often correlate poorly with actual error rates due to miscalibration. A log probability might suggest high confidence numerically, but without calibration, it doesn't guarantee this confidence is justified. Calibrated uncertainties adjust these probabilities to align better with real-world correctness.

\mybenchmark offers a valuable opportunity to develop an external token-level hallucination detection model that produces well-calibrated uncertainty probabilities. Calibrating token probabilities across a large vocabulary is inherently challenging due to the sheer number of possible tokens and the complexity of accurately estimating their individual probabilities. In contrast, a token-level hallucination detection model trained on \mybenchmark only needs to be calibrated at a binary level for each token—simply determining whether a token is hallucinated. This reduction to a binary classification task makes the calibration process more accessible and more intuitive. Moreover, since this model operates externally, it does not interfere with the language model's generation process, allowing us to enhance uncertainty estimation and hallucination detection without impacting the quality of text generation.

\section{Calibration Results and Analysis}
To illustrate that an external hallucination detection model trained on \mybenchmark can enhance the calibration quality of token-level probabilities, we conduct a comprehensive calibration analysis using the Expected Calibration Error (ECE) \citep{Naeini2015ObtainingWC}\footnote{ECE measures the difference between predicted confidence and accuracy over a set of equally-spaced probability bins $B_m$, providing a metric for model calibration. Formally:
\[
\text{ECE} = \sum_{m=1}^M \frac{|B_m|}{n} \left| \text{acc}(B_m) - \text{conf}(B_m) \right|
\]
} and Adaptive Calibration Error (ACE) \citep{nixon2020measuringcalibrationdeeplearning}\footnote{ACE is similar to ECE but uses adaptive binning to ensure each bin has an equal number of samples.} metrics. Calibration measures the agreement between predicted probabilities and actual outcomes, with lower ECE and ACE values indicating better-calibrated models. Tables~\ref{tab:calibraion_instruct}, \ref{tab:calibraion_caption}, and \ref{tab:calibraion_vqa} present the calibration results across three datasets: \textbf{HalLoc-Instruct}, \textbf{HalLoc-Caption}, and \textbf{HalLoc-VQA}.

The models evaluated include:

\begin{itemize}
    \item \textbf{InternVL} \cite{chen2024internvl}: A strong Vision Language Model providing log probabilities.
    \item \textbf{\mymodel (InternVL + VisualBERT)}: Our model trained on \mybenchmark that combines InternVL embeddings with VisualBERT.
    \item \textbf{\mymodel (VisualBERT)}: Our model utilizing only VisualBERT.
\end{itemize}

We report the ECE and ACE values as percentages (\%). Avg represents the macro-average of calibration errors for positive and negative labels\footnote{Due to the natural imbalance in positive labels (hallucinated tokens), it is helpful to analyze each label separately.}.

\subsection{Improvement in Calibration with \mymodel}

Across all datasets, both versions of \mymodel exhibit substantially lower ECE and ACE values than the baseline InternVL model, indicating superior calibration and more reliable probability estimates.


\subsection{Performance Across Different Hallucination Types}

\mymodel versions maintain consistent and robust performance across various hallucination categories, including object, attribute, relationship, and scene. Notably, low ECE and ACE values in \textit{attribute} and \textit{relationship} categories suggest that \mymodel effectively identifies complex hallucinations related to attributes and relationships. This consistent performance indicates that \mymodel can effectively identify subtle and complex hallucinations, enhancing its applicability in diverse scenarios.

\subsection{Impact of Temperature Scaling}
Applying temperature scaling further reduces the ECE and ACE values for both InternVL and \mymodel, enhancing their calibration.
The reduction is more pronounced for InternVL, indicating that it benefits more from calibration techniques but still doesn't match the baseline calibration of \mymodel. This suggests that while temperature scaling is beneficial, \mymodel models inherently possess better calibration than the baseline.


\subsection{Challenge of Detecting Positive Instances}

Our analysis reveals that calibration errors are consistently higher for positive instances—specifically, hallucinated tokens—across all models and datasets. This significant gap between the calibration errors of positive and negative instances highlights a critical obstacle in accurately detecting hallucinated tokens. The challenge is exacerbated by the varying distribution of hallucinated tokens across different tasks (with almost 100\% in \mybenchmarks, 25.37\% in \mybenchmarki, and 5.35\% in \mybenchmarkp for hallucinated samples), making it difficult to train a granular hallucination detection model effectively. Addressing this issue is imperative for advancing the field, and future work must focus on improving the sparse positive labels to enhance detection accuracy and model reliability.

\section{Analyzing the \textit{Grey Area}}
\begin{table}[tb]
\caption{Avg Prob (Original) refers to average hallucination probability before adding Gaussian Noise. Avg Prob ($\sigma$) refers to average hallucination probability after applying Gaussian Noise with different blur intensities ($\sigma$=5, $\sigma$=20).} \centering
    \begin{adjustbox}{max width=\linewidth}
    \large
    \begin{tabular}{l|cccc} \toprule
   & \begin{tabular}{@{}c@{}}Avg Prob  \\ (Original) \end{tabular}  &  \begin{tabular}{@{}c@{}}Avg Prob  \\ ($\sigma$=5) \end{tabular}  &  \begin{tabular}{@{}c@{}}Avg Prob  \\ ($\sigma$=20) \end{tabular}\\ \midrule[0.7pt]
    Overall & 19.70& 19.99& 20.81& \\ 
    \bottomrule
    \end{tabular}
    \end{adjustbox}
    \label{tab:prob_advantage}

\end{table}

In real-world applications, visual data is rarely precise—it often suffers from noise, distortions, or ambiguities caused by factors like poor lighting, motion blur, or environmental interference. We introduce Gaussian noise into images during our experiments to simulate these imperfections, aiming to evaluate how \mymodel handles such visual noise. This approach is critical as it probes the grey area where the model’s interpretations oscillate.

In our framework, object hallucination includes cases where descriptions—such as attributes and relationships—pertain to nonexistent objects. To explore this, we selectively apply Gaussian noise: to the object bounding boxes pertaining to truthful objects(including those describing attributes and relationships). For scene tokens, Gaussian noise is added to the entire image.

The results, summarized in Table \ref{tab:prob_advantage}, reveal a gradual increase in the likelihood of hallucinated tokens as image noise intensifies, rather than an abrupt shift from non-hallucination to hallucination. This nuanced progression underscores the limitations of binary classification in capturing such subtle transitions. Consequently, traditional binary metrics may fail to reflect the model's performance under varying noise conditions adequately. To address this, incorporating probabilistic or spectrum-based evaluation methods could provide a more detailed understanding of the model's behavior in the face of visual uncertainty.
\begin{table*}[h!]
\begin{adjustbox}{max width=\textwidth} \centering
\begin{tabular}{llccccccccccc}
\toprule
\multirow{2}{*}{\begin{tabular}[c]{@{}l@{}}Probability\\ Model\end{tabular}} & \multirow{2}{*}{\begin{tabular}[c]{@{}c@{}}Calibration\\Techniques\end{tabular}} & \multirow{2}{*}{Label} & \multicolumn{2}{c}{total} & \multicolumn{2}{c}{object} & \multicolumn{2}{c}{attribute} & \multicolumn{2}{c}{relationship} & \multicolumn{2}{c}{scene} \\
\cmidrule[0.5pt](rl){4-5} \cmidrule[0.5pt](rl){6-7} \cmidrule[0.5pt](rl){8-9} \cmidrule[0.5pt](rl){10-11} \cmidrule[0.5pt](rl){12-13}
 &  &  & ECE & ACE & ECE & ACE & ECE & ACE & ECE & ACE & ECE & ACE \\ \midrule[0.7pt]
\multirow{6}{*}{InternVL} & \multirow{3}{*}{original} & pos & 73.72 & 67.21 & - & - & - & - & - & - & - & - \\
 &  & neg & 91.60 & 78.17 & - & - & - & - & - & - & - & - \\
 &  & avg & 82.66 & 72.69 & - & - & - & - & - & - & - & - \\
 \cmidrule[0.5pt](rl){4-13}
 & \multirow{3}{*}{+TS} & pos & 39.25& 38.67 &  & - & - & - & - & - & - & -  \\
 &  & neg & 53.31 &  44.47 & - & - & - & - & - & - & - & - \\
 &  & avg & 46.28 & 41.57  & - & - & - & - & - & - & - & - \\ \midrule
\multirow{6}{*}{\begin{tabular}[c]{@{}l@{}}HalLocalizer\\ (InternVL + \\ VisualBERT)\end{tabular}} & \multirow{3}{*}{original} & pos & - & - & 20.92 & 20.80 & 0.86 & 1.37 & 28.71 & 28.48 & 3.23 & 3.02 \\
 &  & neg & - & - & 21.98 & 22.03 & 12.13 & 12.19 & 11.50 & 11.53 & 1.67 & 1.75 \\
 &  & avg & - & - & 21.45 & 21.41 & 6.50 & 6.78 & 20.11 & 20.00 & 2.45 & 2.38  \\ 
 \cmidrule[0.5pt](rl){4-13}
 & \multirow{3}{*}{+TS} & pos & - & - & 18.80 & 18.36 & 6.80 & 7.00 & 24.83 & 24.66 & 2.23 & 2.20 \\
 &  & neg & - & - & 20.64 & 20.87 & 8.86 & 11.25 & 9.43 & 9.82 & 1.22 & 1.69 \\
 &  & avg & - & - & 19.72 & 19.62 & 7.83 & 9.12 & 17.13 & 17.24 & 1.73 & 1.95  \\ \midrule
\multirow{6}{*}{\begin{tabular}[c]{@{}l@{}}HalLocalizer\\ (VisualBERT)\end{tabular}} & \multirow{3}{*}{original} & pos & - & - & 30.97 & 30.68 & 17.96 & 18.17 & 27.51 & 27.64 & 20.42 & 19.20 \\
 &  & neg & - & - & 5.36 & 5.41 & 6.12 & 6.13 & 7.10 & 7.16 & 0.96 & 0.97 \\
 &  & avg & - & - & 18.16 & 18.05 & 12.04 & 12.15 & 17.30 & 17.40 & 10.69 & 10.08  \\ 
 \cmidrule[0.5pt](rl){4-13}
 & \multirow{3}{*}{+TS} & pos & - & - & 27.34 & 27.29 & 13.15 & 13.89 & 24.03 & 23.99 & 17.72 & 16.11 \\
 &  & neg & - & - & 4.28 & 4.66 & 3.45 & 5.30 & 5.77 & 6.15 & 0.68 & 1.02 \\
 &  & avg & - & - & 15.81 & 15.97 & 8.30 & 9.60 & 14.90 & 15.07 & 9.20 & 8.56 \\
 \bottomrule
\end{tabular}
\end{adjustbox}
\caption{Probability Calibration of HalLocalizer on \textbf{HalLoc-VQA}. TS stands for Temperature Scaling.}
\label{tab:calibraion_vqa}
\end{table*}
\begin{table*}[]
\begin{adjustbox}{max width=\textwidth} \centering
\begin{tabular}{llccccccccccc}
\toprule
\multirow{2}{*}{\begin{tabular}[c]{@{}l@{}}Probability\\ Model\end{tabular}} & \multirow{2}{*}{\begin{tabular}[c]{@{}l@{}}Calibration\\ Techniques\end{tabular}} & \multirow{2}{*}{Label} & \multicolumn{2}{c}{total} & \multicolumn{2}{c}{object} & \multicolumn{2}{c}{attribute} & \multicolumn{2}{c}{relationship} & \multicolumn{2}{c}{scene} \\
\cmidrule[0.5pt](rl){4-5} \cmidrule[0.5pt](rl){6-7} \cmidrule[0.5pt](rl){8-9} \cmidrule[0.5pt](rl){10-11} \cmidrule[0.5pt](rl){12-13}
 &  &  & ECE & ACE & ECE & ACE & ECE & ACE & ECE & ACE & ECE & ACE \\ \midrule[0.7pt]
\multirow{6}{*}{InternVL} & \multirow{3}{*}{original} & pos & 64.44 &  64.44& - & - & - & - & - & - & - & - \\
 &  & neg &  78.90 &  72.71& - & - & - & - & - & - & - & - \\
 &  & avg & 71.67 & 68.57 & - & - & - & - & - & - & - & - \\
 \cmidrule[0.5pt](rl){4-13}
 & \multirow{3}{*}{+TS} & pos & 31.79 &  32.01& - & - & - & - & - & - & - & - \\
 &  & neg & 43.08 &  37.37& - & - & - & - & - & - & - & - \\
 &  & avg & 37.44 & 34.66 & - & - & - & - & - & - & - & - \\ \midrule
\multirow{6}{*}{\begin{tabular}[c]{@{}l@{}}HalLocalizer\\ (InternVL + \\ VisualBERT)\end{tabular}} & \multirow{3}{*}{original} & pos & - & - & 22.93 & 22.95 & 1.34 & 1.19 & 8.83 & 8.61 & 1.65 & 1.78 \\
 &  & neg & - & - & 0.11 & 0.13 & 1.95 & 2.00 & 5.83 & 5.89 & 0.29 & 0.36 \\
 &  & avg & - & - & 11.52 & 11.54 & 1.65 & 1.59 & 7.33 & 7.25 & 0.97 & 1.07  \\  \cmidrule[0.5pt](rl){4-13}
 & \multirow{3}{*}{+TS} & pos & - & - & 21.75 & 21.70 & 2.56 & 2.41 & 7.55 & 7.34 & 1.11 & 2.00 \\
 &  & neg & - & - & 0.33 & 0.34 & 0.97 & 2.39 & 5.51 & 5.73 & 0.26 & 0.48 \\
 &  & avg & - & - & 11.04 & 11.02 & 1.77 & 2.40 & 6.53 & 6.54 & 0.69 & 1.24  \\ \midrule
\multirow{6}{*}{\begin{tabular}[c]{@{}l@{}}HalLocalizer\\ (VisualBERT)\end{tabular}} & \multirow{3}{*}{original} & pos & - & - & 20.42 & 20.40 & 0.78 & 0.44 & 8.94 & 8.95 & 2.43 & 2.09 \\
 &  & neg & - & - & 0.19 & 0.16 & 1.73 & 1.76 & 6.32 & 6.33 & 0.23 & 0.28 \\
 &  & avg & - & - & 10.31 & 10.28 & 1.25 & 1.10 & 7.63 & 7.64 & 1.33 & 1.19  \\ \cmidrule[0.5pt](rl){4-13}
 & \multirow{3}{*}{+TS} & pos & - & - & 19.69 & 19.63 & 1.07 & 0.85 & 2.31 & 2.16 & 3.87 & 3.77 \\
 &  & neg & - & - & 0.14 & 0.18 & 1.48 & 1.79 & 4.16 & 5.83 & 0.24 & 0.36 \\
 &  & avg & - & - & 9.92 & 9.90 & 1.27 & 1.32 & 3.24 & 4.00 & 2.06 & 2.06 \\ \bottomrule
\end{tabular}
\end{adjustbox}
\caption{Probability Calibration of HalLocalizer on \textbf{HalLoc-Instruct}. TS stands for Temperature Scaling.}
\label{tab:calibraion_instruct}
\end{table*}

\begin{table*}[]
\begin{adjustbox}{max width=\textwidth} \centering
\begin{tabular}{llccccccccccc}
\toprule
\multirow{2}{*}{\begin{tabular}[c]{@{}l@{}}Probability\\ Model\end{tabular}} & \multirow{2}{*}{\begin{tabular}[c]{@{}l@{}}Calibration\\ Techniques\end{tabular}} & \multirow{2}{*}{Label} & \multicolumn{2}{c}{total} & \multicolumn{2}{c}{object} & \multicolumn{2}{c}{attribute} & \multicolumn{2}{c}{relationship} & \multicolumn{2}{c}{scene} \\
\cmidrule[0.5pt](rl){4-5} \cmidrule[0.5pt](rl){6-7} \cmidrule[0.5pt](rl){8-9} \cmidrule[0.5pt](rl){10-11} \cmidrule[0.5pt](rl){12-13}
 &  &  & ECE & ACE & ECE & ACE & ECE & ACE & ECE & ACE & ECE & ACE \\ \midrule[0.7pt]
\multirow{6}{*}{InternVL} & \multirow{3}{*}{original} & pos & 46.47 & 46.47 & - & - & - & - & - & - & - & - \\
 &  & neg & 64.86 & 46.61 & - & - & - & - & - & - & - & - \\
 &  & avg & 55.66 & 46.54& - & - & - & - & - & - & - & - \\  \cmidrule[0.5pt](rl){4-13}
 & \multirow{3}{*}{+TS} & pos & 16.51 & 15.57 & - & - & - & - & - & - & - & - \\
 &  & neg & 31.63 & 14.16 & - & - & - & - & - & - & - & - \\
 &  & avg & 24.07 & 14.87& - & - & - & - & - & - & - & - \\ \midrule
\multirow{6}{*}{\begin{tabular}[c]{@{}l@{}}HalLocalizer\\ (InternVL + \\ VisualBERT)\end{tabular}} & \multirow{3}{*}{original} & pos & - & - & 37.29 & 37.13 & 51.34 & 50.80 & 44.07 & 49.00 & 81.98 & 81.88 \\
 &  & neg & - & - & 0.04 & 0.08 & 0.57 & 0.54 & 0.17 & 0.13 & 0.10 & 0.10 \\
 &  & avg & - & - & 18.66 & 18.61 & 25.96 & 25.67 & 22.12 & 24.57 & 41.04 & 40.99 \\ \cmidrule[0.5pt](rl){4-13}
 & \multirow{3}{*}{+TS} & pos & - & - & 34.34 & 34.32 & 51.22 & 50.68 & 40.68 & 45.81 & 79.86 & 79.76 \\
 &  & neg & - & - & 0.29 & 0.27 & 0.59 & 0.56 & 0.63 & 0.60 & 0.39 & 0.39 \\
 &  & avg & - & - &17.32 & 17.30 & 25.91 & 25.62 & 20.66 & 23.21 & 40.12 & 40.08  \\ \midrule
\multirow{6}{*}{\begin{tabular}[c]{@{}l@{}}HalLocalizer\\ (VisualBERT)\end{tabular}} & \multirow{3}{*}{original} & pos &  &  & 21.61 & 21.41 & 31.30 & 31.30 & 21.31 & 19.46 & 22.02 & 24.05 \\
 &  & neg & - & - & 0.14 & 0.18 & 0.11 & 0.16 & 0.20 & 0.24 & 0.02 & 0.02 \\
 &  & avg & - & - & 10.88 & 10.79 & 15.71 & 15.73 & 10.75 & 9.85 & 11.02 & 12.04  \\  \cmidrule[0.5pt](rl){4-13}
 & \multirow{3}{*}{+TS} & pos &  &  & 17.03 & 16.95 & 28.98 & 28.99 & 18.77 & 17.22 & 20.45 & 21.46 \\
 &  & neg & - & - & 0.45 & 0.40 & 0.24 & 0.21 & 0.30 & 0.26 & 0.17 & 0.16 \\
 &  & avg & - & - & 8.74 & 8.67 & 14.61 & 14.60 & 9.54 & 8.74 & 10.31 & 10.81 \\ \bottomrule
\end{tabular}
\end{adjustbox}
\caption{Probability Calibration of HalLocalizer on \textbf{HalLoc-Caption}. TS stands for Temperature Scaling.}
\label{tab:calibraion_caption}
\end{table*}
\clearpage
\clearpage
\section{Illustrative Examples of \mybenchmark}
\label{examples}
\begin{figure}[b!]
  \centering
  \includegraphics[width=\linewidth]{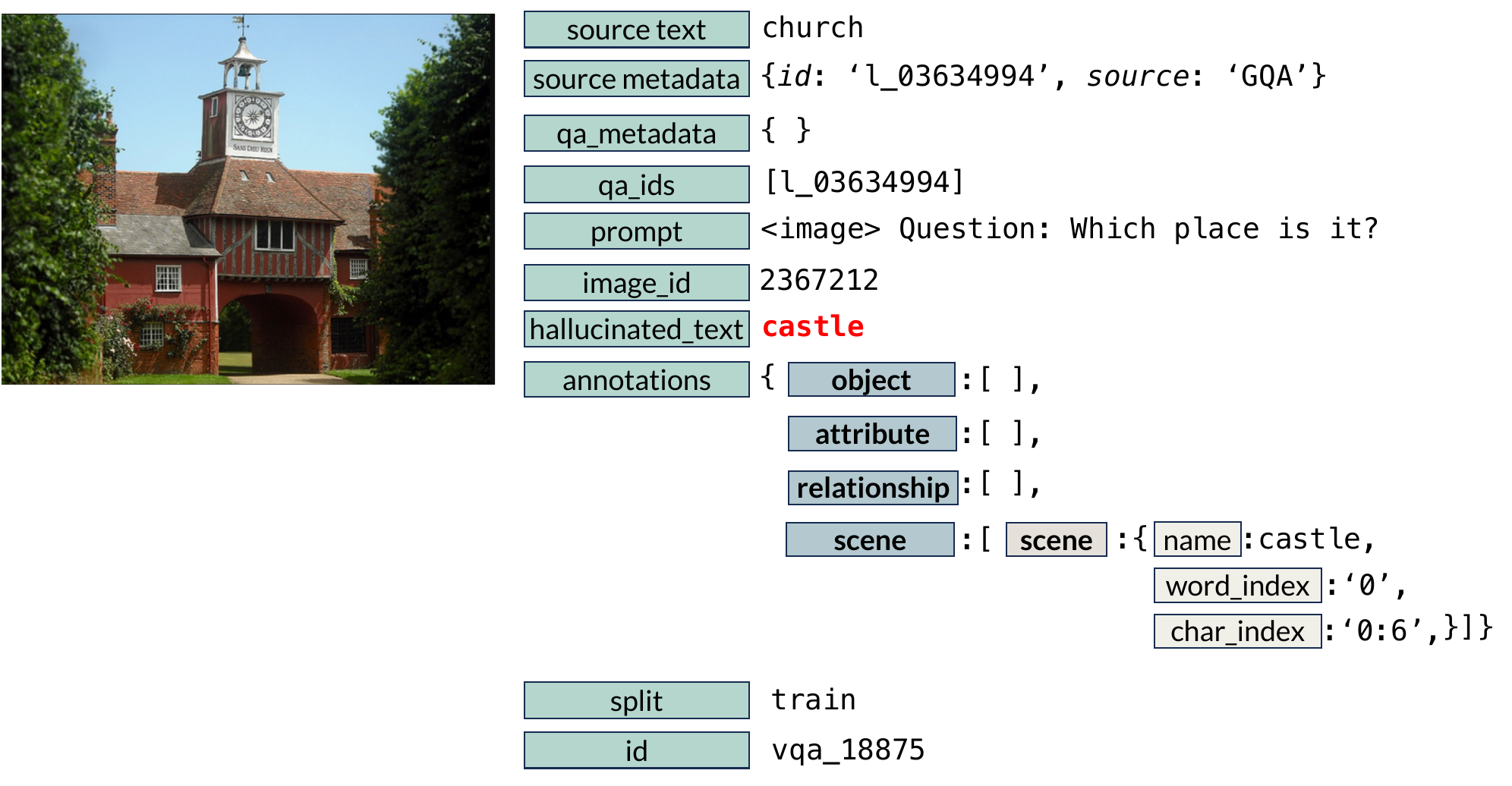}
  \caption{Example of an entry in \mybenchmarks of hallucination type \textbf{scene}.
  }
  \label{fig:halloc_vqa_ex}
\end{figure}

\begin{figure}[b!]
  \centering
  \includegraphics[width=\linewidth]{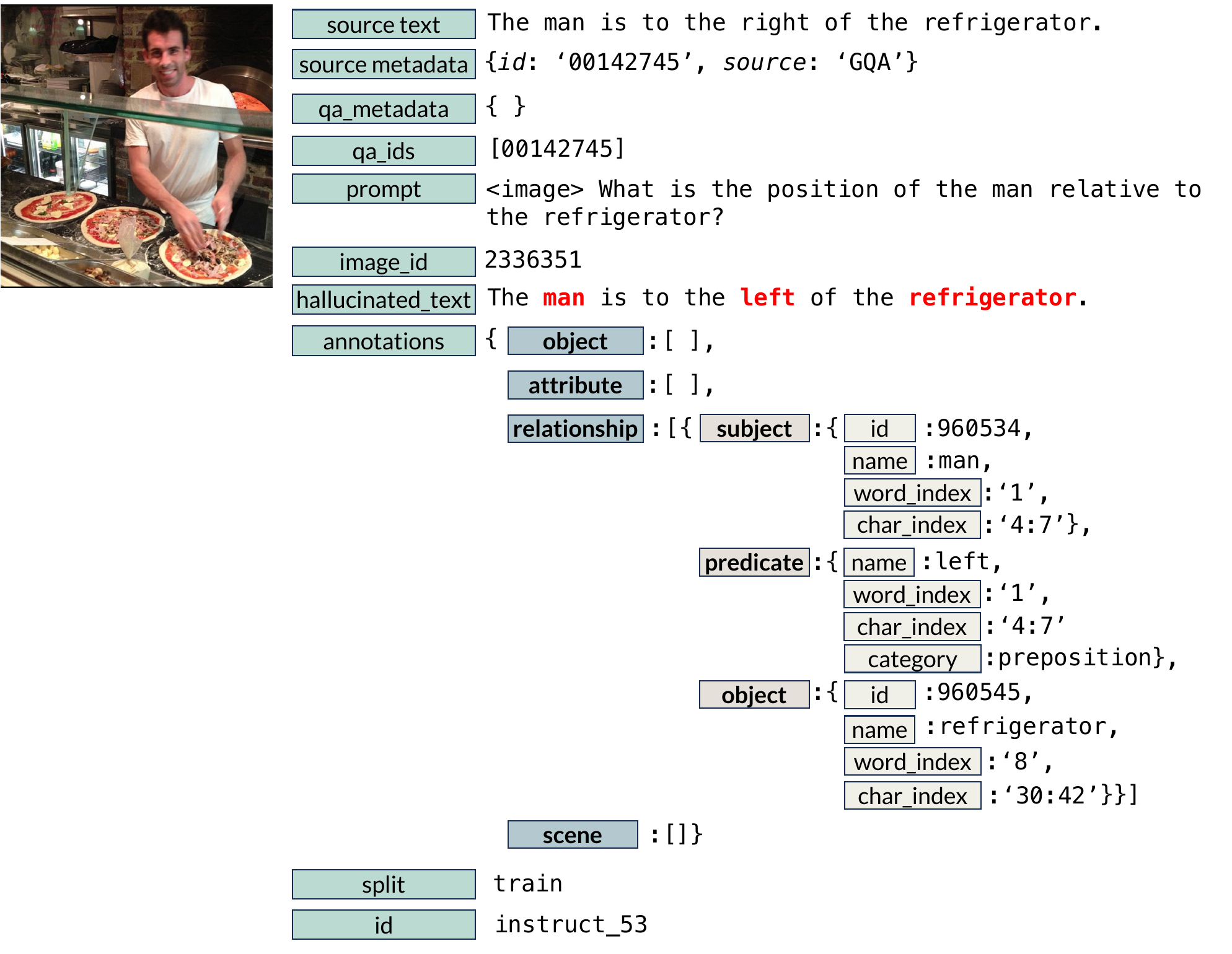}
  \caption{Example of an entry in \mybenchmarki of hallucination type \textbf{relationship}.
  }
  \label{fig:halloc_instruct_ex}
\end{figure}

\begin{figure}[b!]
  \centering
  \includegraphics[width=\linewidth]{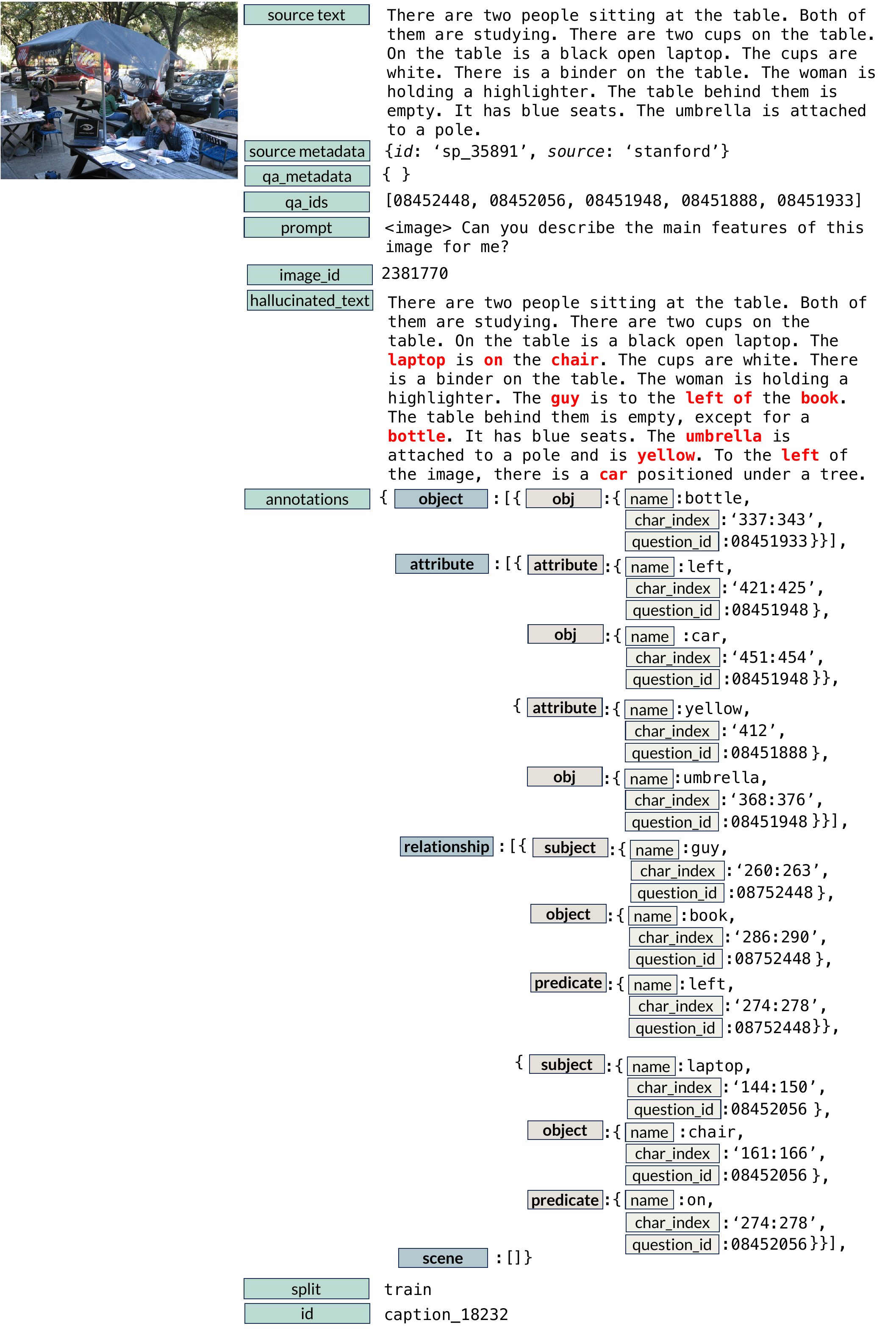}
  \caption{Example of an entry in \mybenchmarkp comprising of multiple types of hallucination; \textbf{object}, \textbf{attribute}, \textbf{relationship}.
  }
  \label{fig:halloc_cap_ex}
\end{figure}

\begin{figure}[b!]
    \centering
    \includegraphics[width=\linewidth]{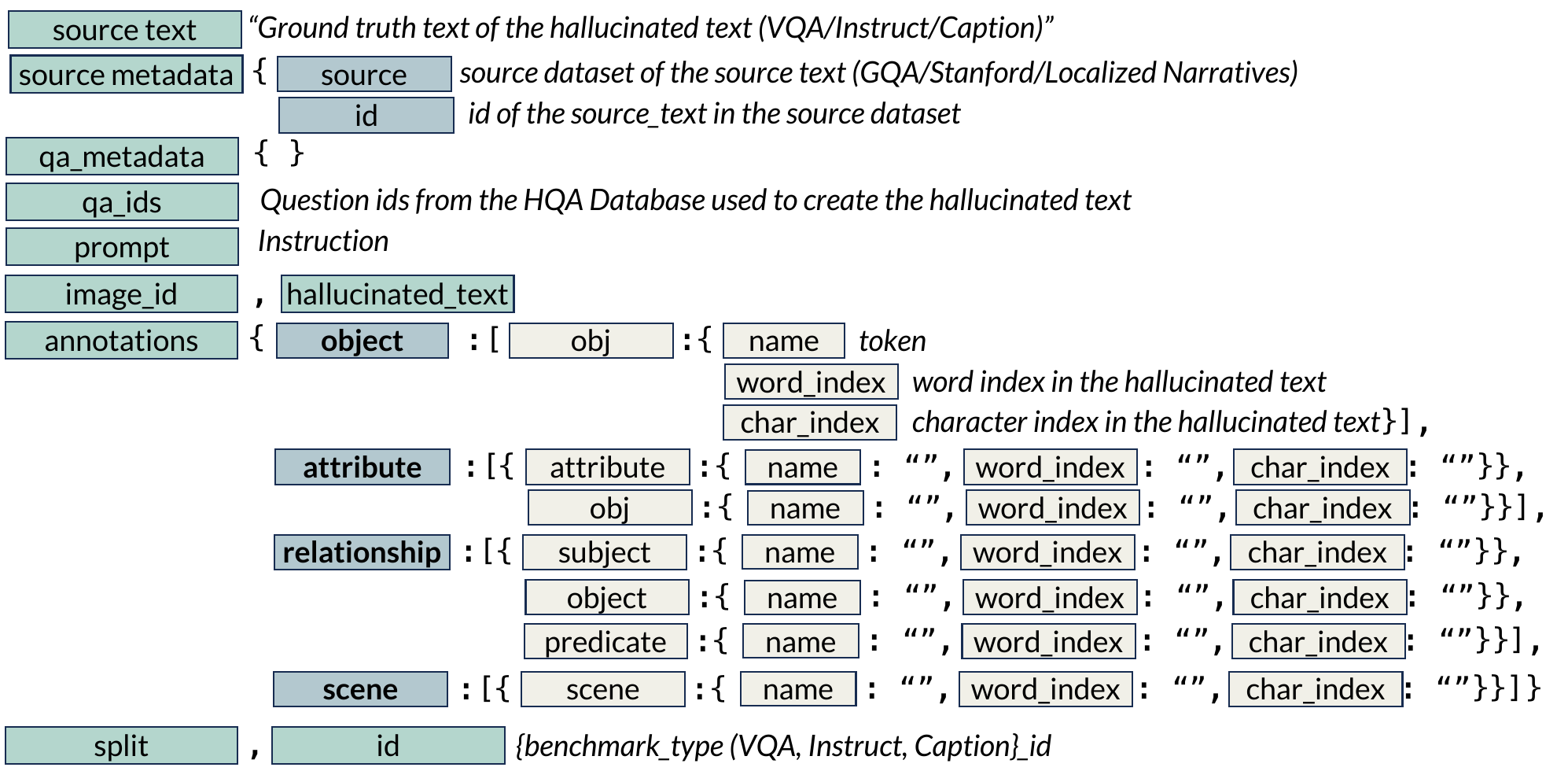}
    \caption{Annotation format of \mybenchmark}
    \label{fig:halloc_annotation_format}
\end{figure}



We provide examples of data points in \mybenchmark\ for each of \mybenchmarks, \mybenchmarki, and \mybenchmarkp in Figures \ref{fig:halloc_vqa_ex}, \ref{fig:halloc_instruct_ex}, and \ref{fig:halloc_cap_ex}.
The general annotation format of \mybenchmark\ is illustrated in Figure \ref{fig:halloc_annotation_format}.

\clearpage
\section{\mybenchmark\ Instruction Details}
\label{halloc_bench_details}
\label{instructions}
\begin{figure}[tbh]
  \centering
  \includegraphics[width=\linewidth]{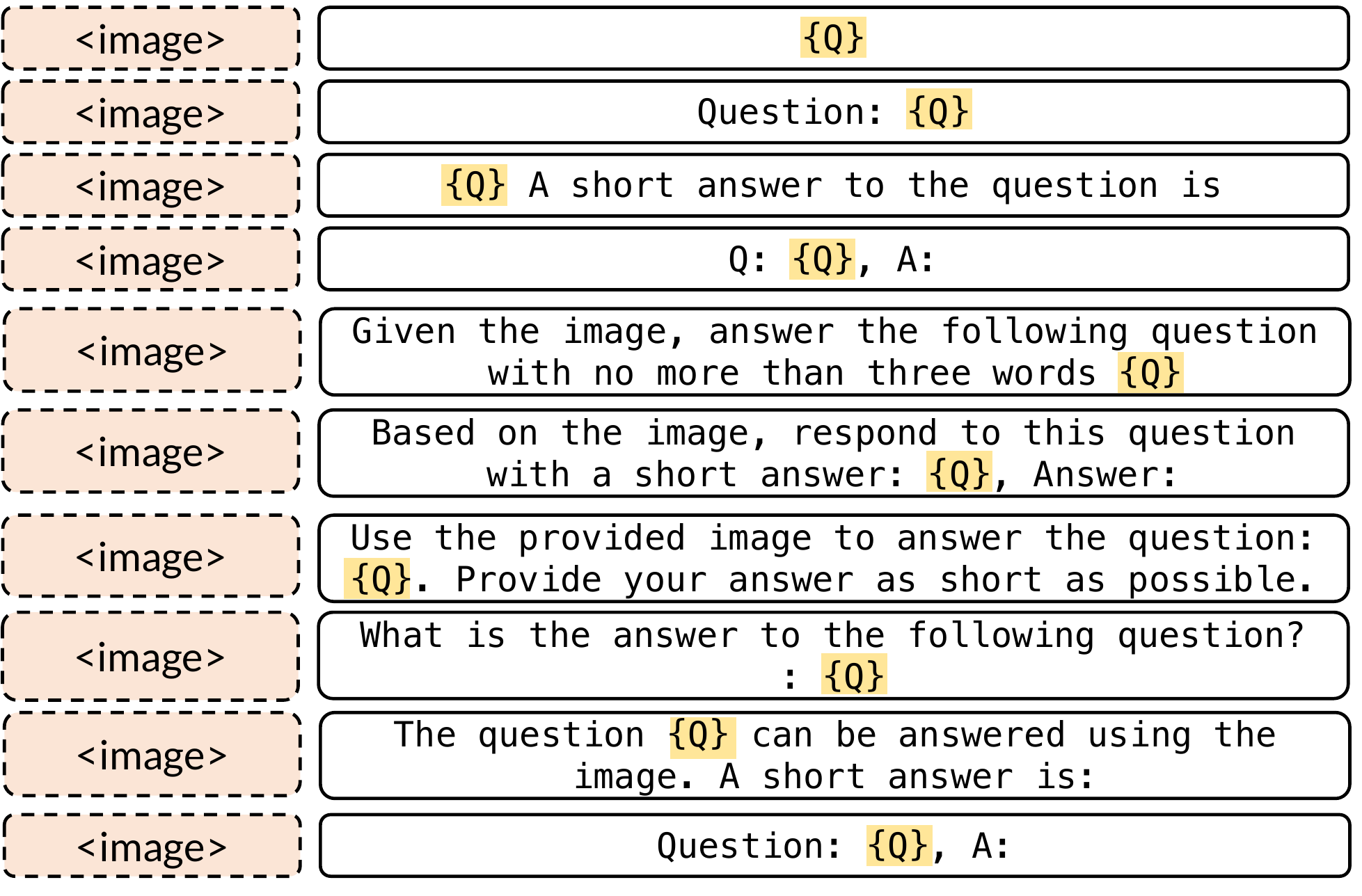}
  \caption{Examples of instructions for model responses in \mybenchmarks. Q is a placeholder for the visual question.}
  \label{fig:instructions_vqa}
\end{figure}

\begin{figure}[tbh]
  \centering
  \includegraphics[width=\linewidth]{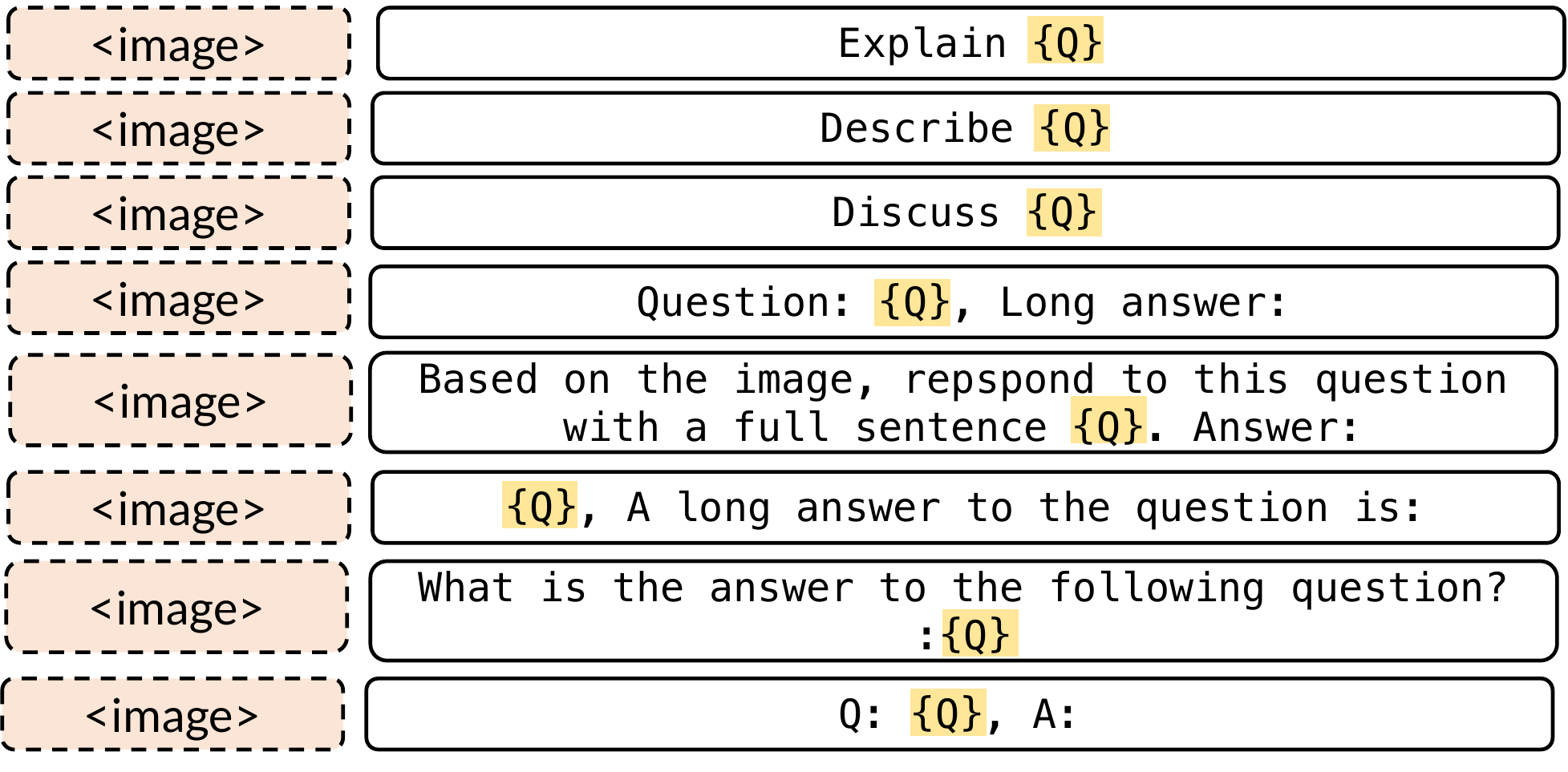}
  \caption{Examples of instructions for model responses in \mybenchmarki. Q is a placeholder for the visual question.}
  \label{fig:instructions_instruct}
\end{figure}

\begin{figure}[tbh]
  \includegraphics[width=\linewidth]{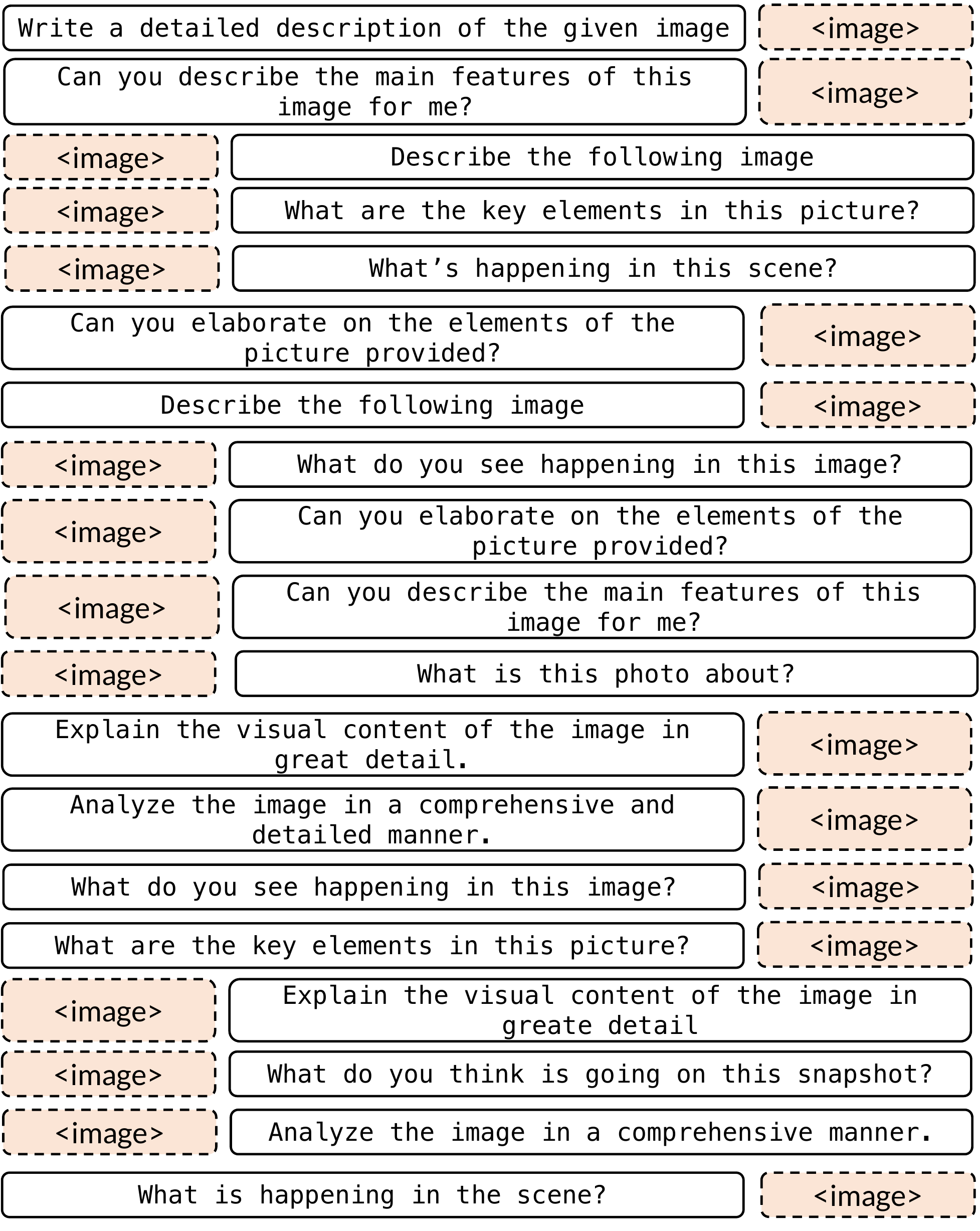}
  \caption{Examples of instructions for model responses in \mybenchmarkp.}
  \label{fig:instructions_cap}
\end{figure}

Each entry of \mybenchmark\ consists of an \{\textbf{Instruction}\}-\{\textbf{Response}\} pair.
We show the details of the templates we used for the instructions in \mybenchmarks in Figure \ref{fig:instructions_vqa}, \mybenchmarki in Figure \ref{fig:instructions_instruct} and \mybenchmarkp in Figure \ref{fig:instructions_cap}.

\clearpage
\section{HQA Injection Pipeline Details}

\subsection{Annotating QA pairs}
\label{prompts_for_qa}
\begin{figure}[h!]
  \centering
  \includegraphics[width=\linewidth]{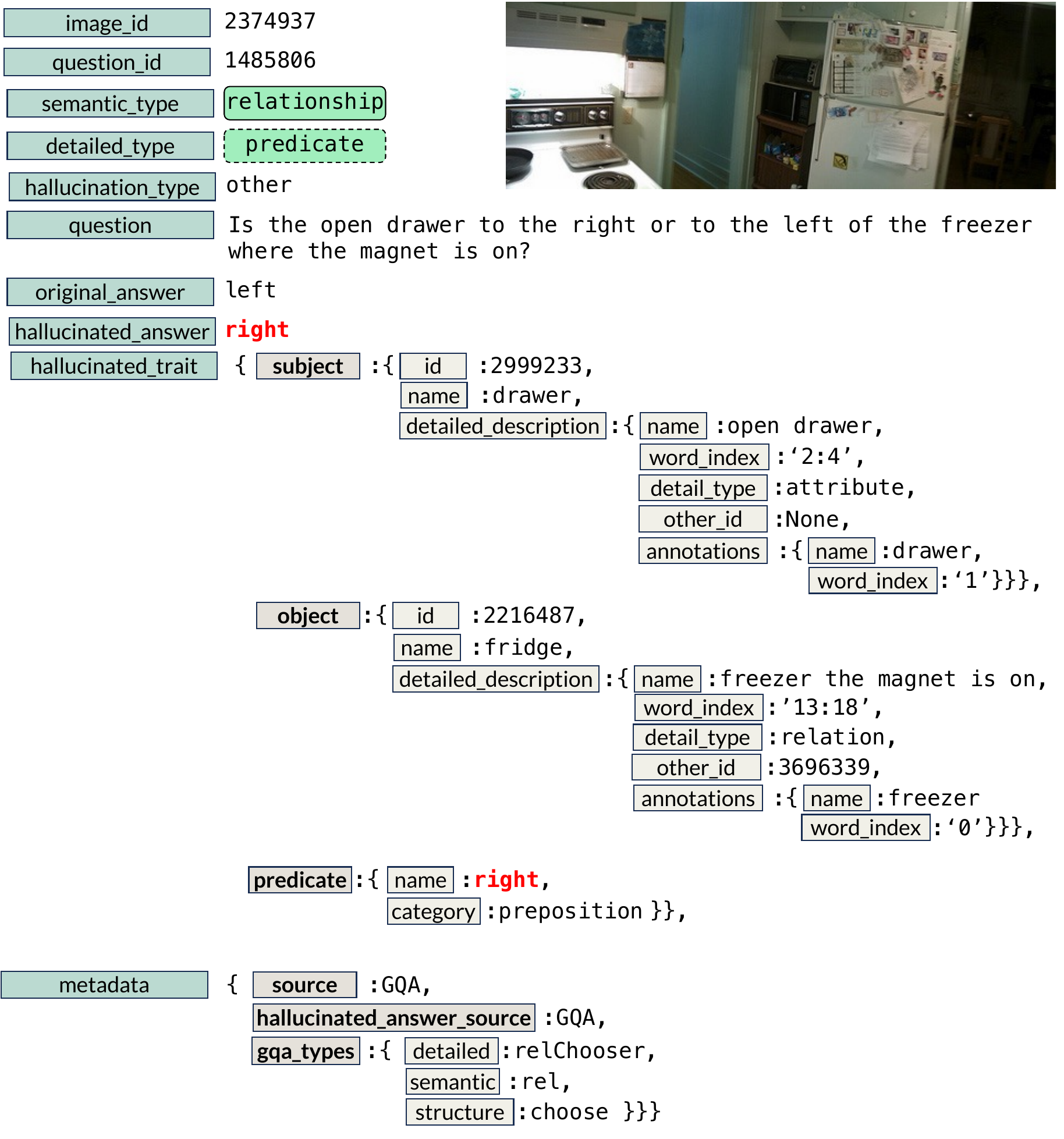}
  \caption{Example of how each component in a single visual QA pair in the GQA dataset is annotated with hallucinating question and answer in \mybenchmark's HQA Database. 
  }
  \label{fig:annotating_qa}
\end{figure}

\begin{figure}[h!]
  \centering
  \includegraphics[width=\linewidth]{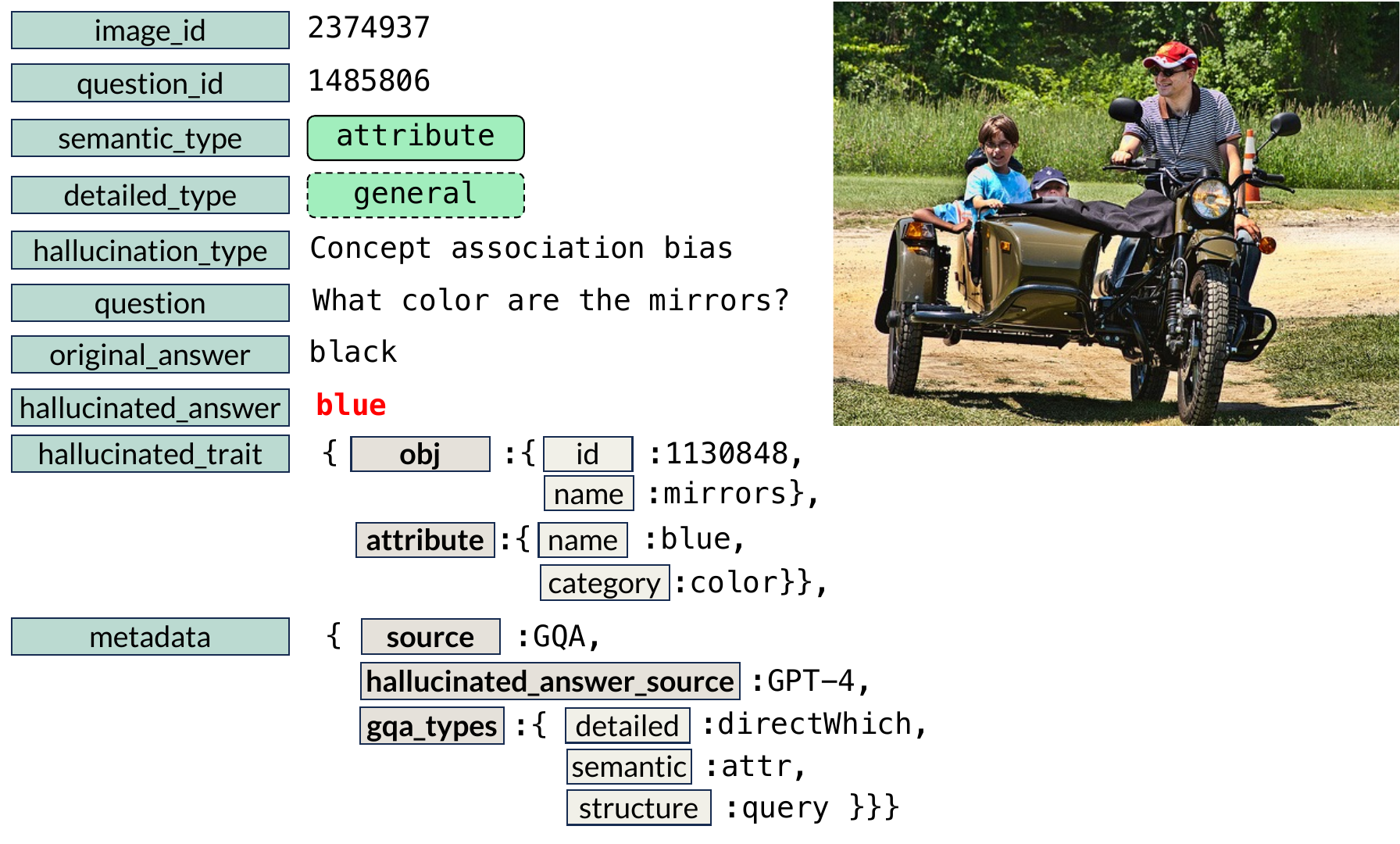}
  \caption{Example of how each component in a single visual QA pair in the GQA dataset is annotated with hallucinating question and answer in \mybenchmark's HQA Database.
  }
  \label{fig:annotating_qa_cab}
\end{figure}

\mybenchmark\ utilizes the GQA dataset \citep{hudson2019gqa} as the foundation for its questions and answers. 
In particular, we take advantage of the questions and scene graphs provided by GQA to achieve detailed annotations of the components within each question and answer. 
Figures \ref{fig:annotating_qa} and \ref{fig:annotating_qa_cab} show examples of how we annotate a GQA question and answer to save them in the HQA Database. The hallucinated answer in Figure \ref{fig:annotating_qa} is directly derived from the GQA question (the other choice). The hallucinated answer in Figure \ref{fig:annotating_qa_cab} comes from crafted hallucinated candidates. 

\subsection{Crafting Hallucinated Answers}
\label{crafting_prompts}

\begin{figure*}[t!]
  \centering
  \includegraphics[width=0.8\textwidth]{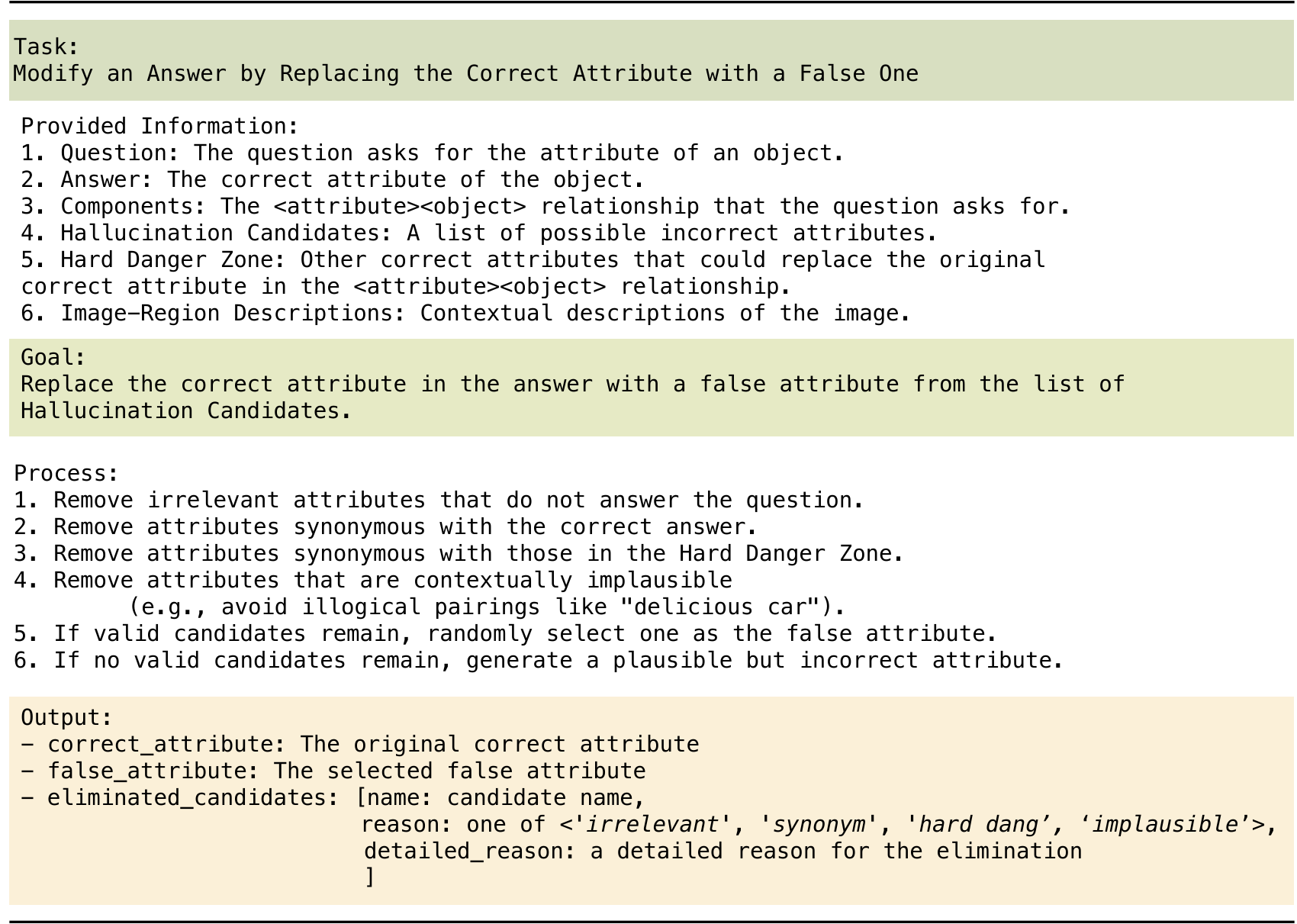}
  \caption{Prompt used to craft hallucinated answers for hallucination type attribute.
  }
  \label{fig:crafting_attr}
\end{figure*}

\begin{figure*}[t!]
  \centering
  \includegraphics[width=0.8\linewidth]{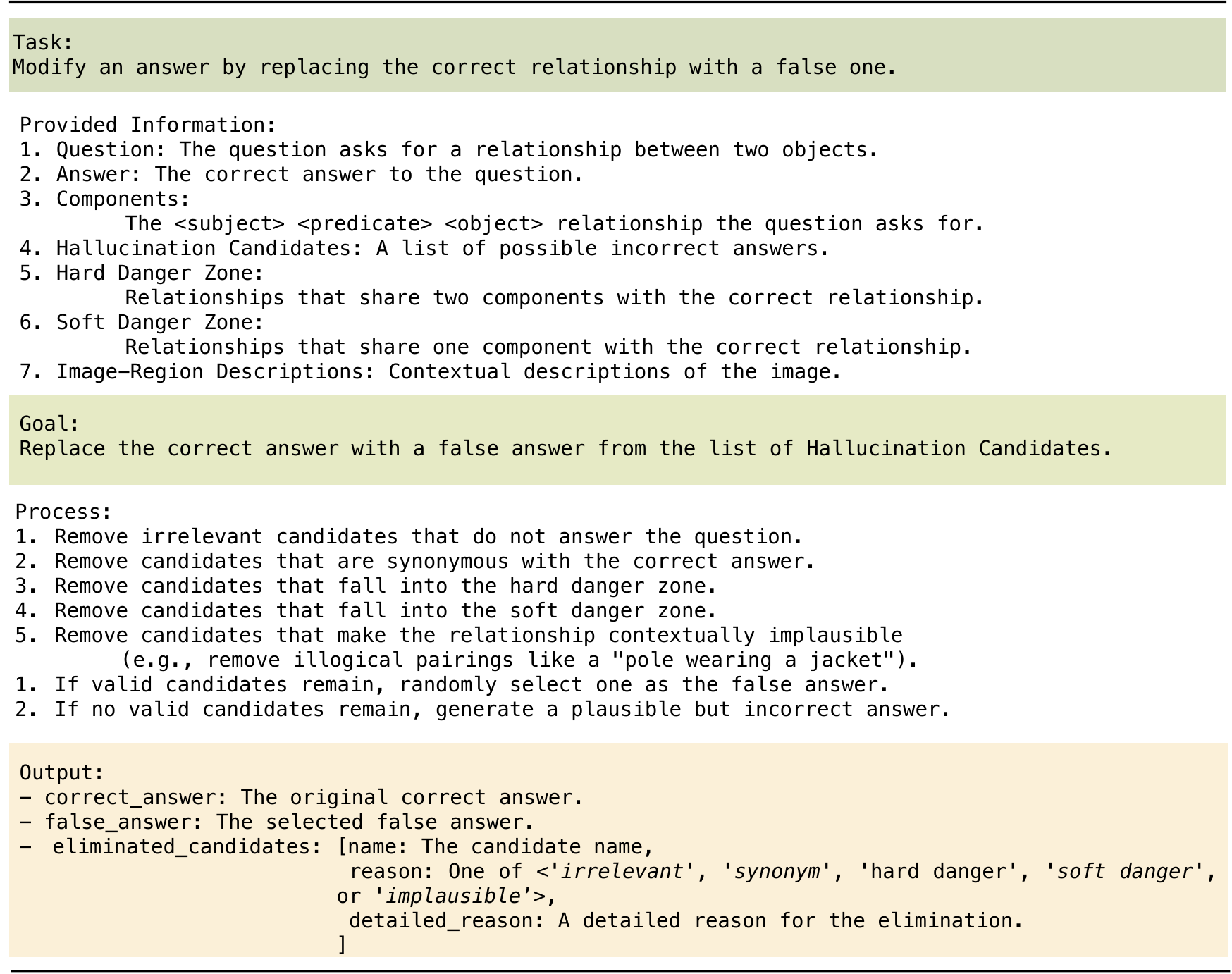}
  \caption{Prompt used to craft hallucinated answers for hallucination type relationship.
  }
  \label{fig:crafting_rel}
\end{figure*}

After we generate a set of hallucinated answer candidates that reflect common causes and patterns of hallucinations in Large Vision-Language Models (LVLMs), we use GPT-4 \citep{openai2024gpt4} to choose a hallucinated answer from these candidates. Specifically, GPT-4 is employed to assess hallucinated answer candidates for attribute and relationship questions, which demand advanced reasoning to avoid generating nonsensical or overly similar responses. 
Figures \ref{fig:crafting_attr} and \ref{fig:crafting_rel} show the prompts used to evaluate the hallucinated answers. 
In practice, each prompt is associated with in-context examples to demonstrate how the process works.

\subsection{Determining Injection Points}
\label{sec:injecting_prompts}
We pose the question to the paragraph to identify where to inject the hallucinated answers. 
Specifically, subsections of the paragraph that share similarities with the hallucinated elements the question seeks become ideal candidates for injection. 
A brief reminder that the hallucinated elements for each question type are:

Object: \(\texttt{<obj>}\)

Attribute: \(\texttt{<attr><obj>}\)

Relationship: \(\texttt{<obj1><rel><obj2>}\)

Scene: \(\texttt{<sce>}\)

\begin{figure*}[t!]
  \centering
  \includegraphics[width=0.8\linewidth]{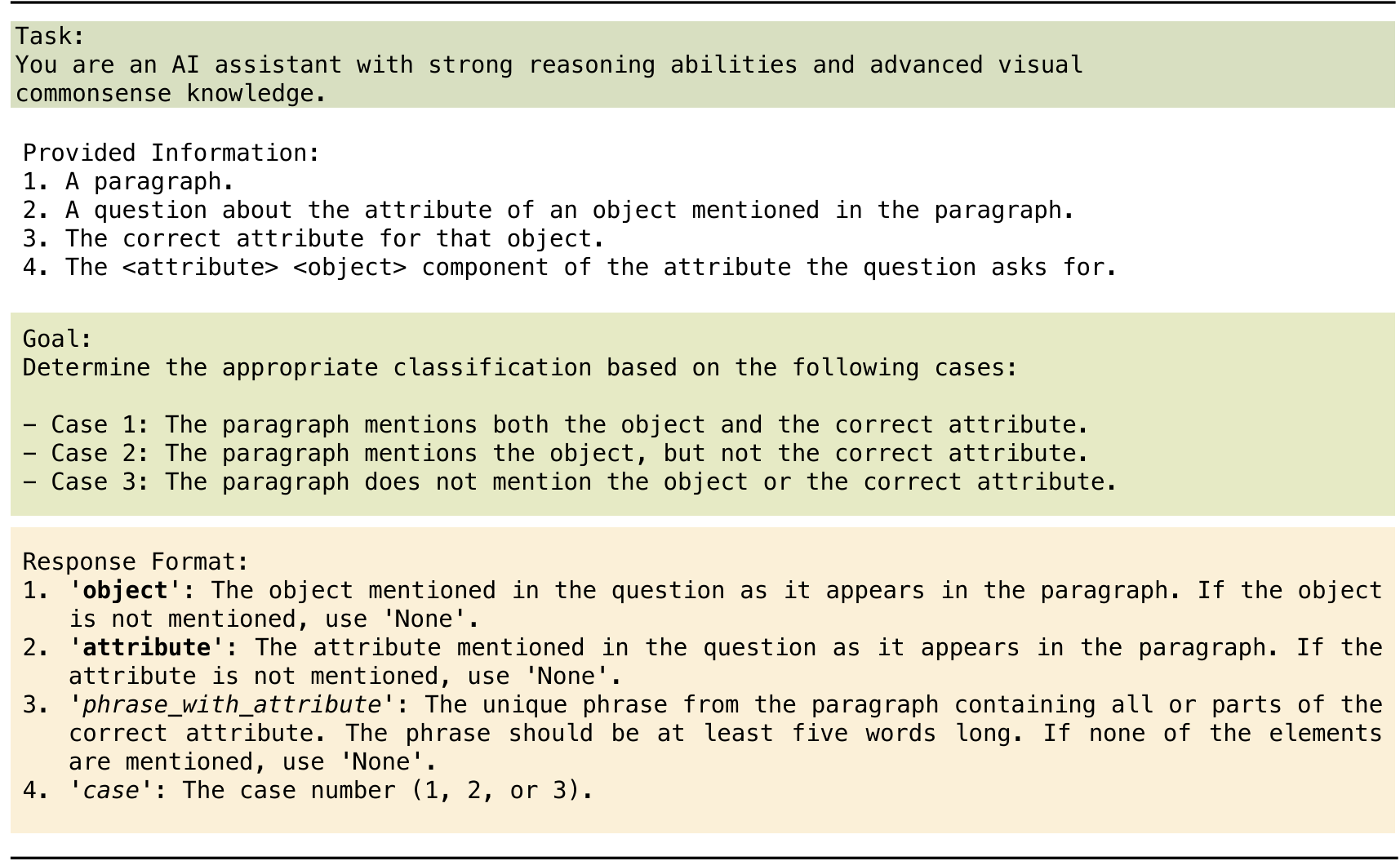}
  \caption{Prompt used to determine the injection points of the hallucinated answers to source texts for hallucination type attribute.
  }
  \label{fig:det_inject_att}
\end{figure*}

\begin{figure*}[t!]
  \centering
  \includegraphics[width=0.8\linewidth]{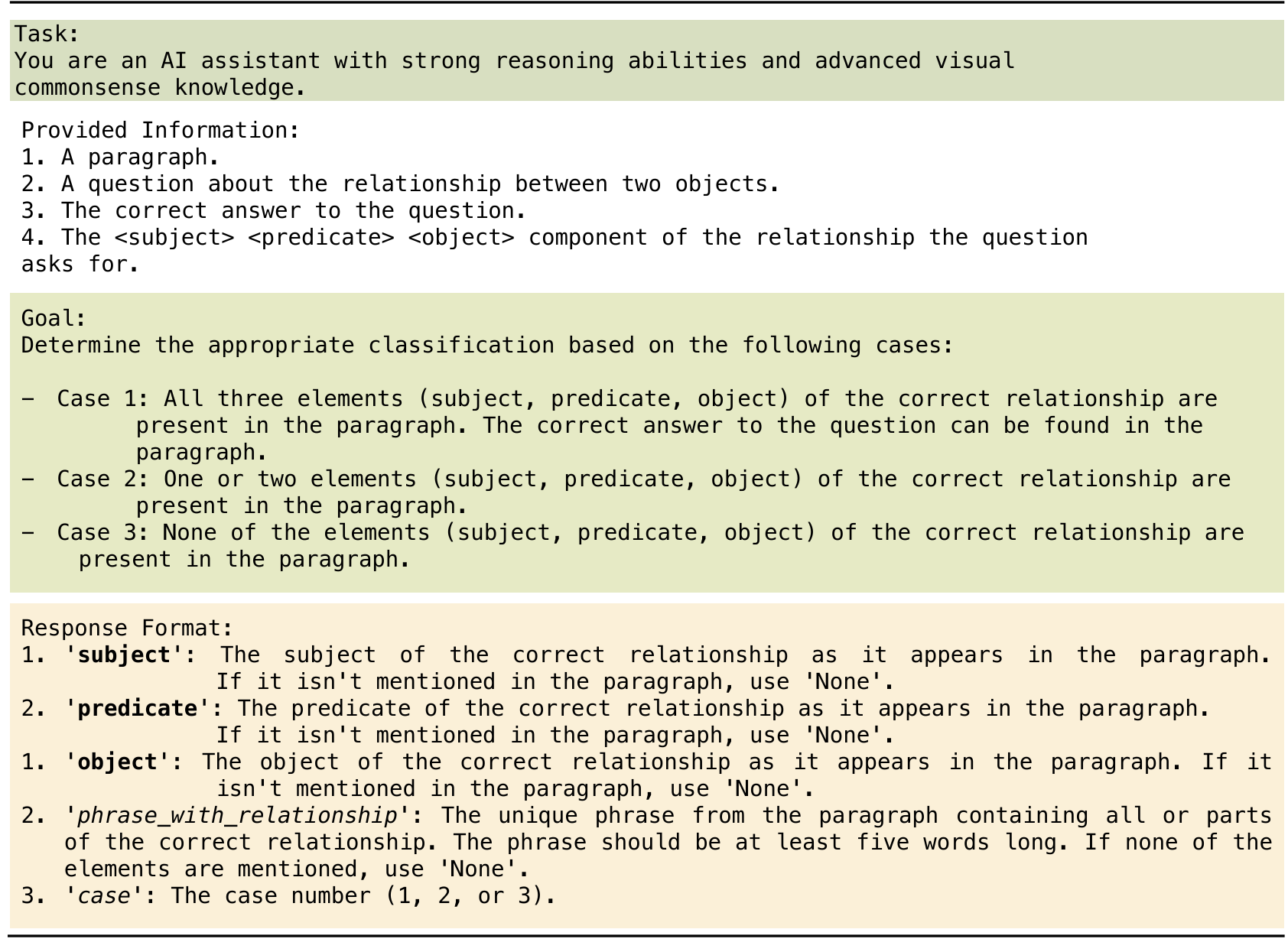}
  \caption{Prompt used to determine the injection points of the hallucinated answers to source texts for hallucination type relationship.
  }
  \label{fig:det_inject_rel}
\end{figure*}

\begin{figure*}[t!]
  \centering
  \includegraphics[width=0.8\linewidth]{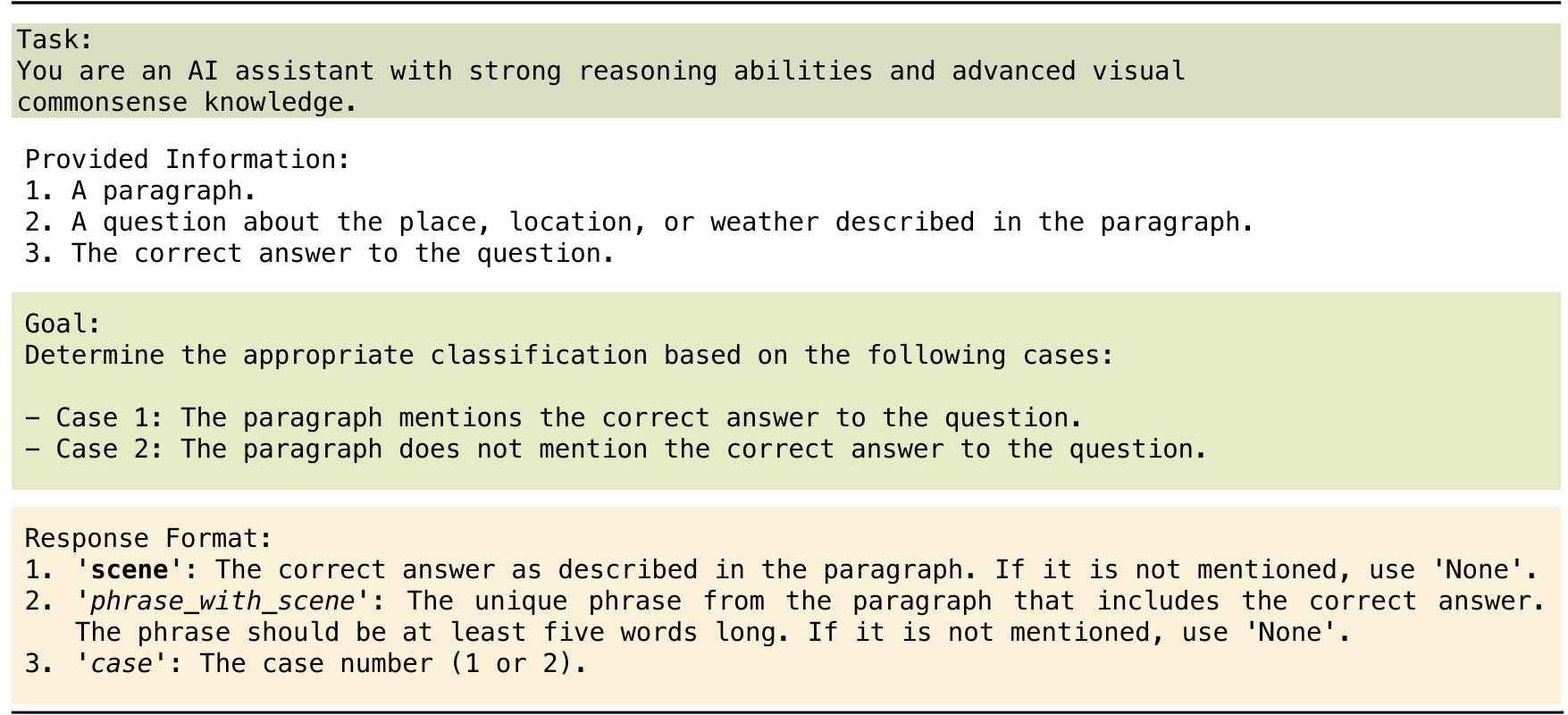}
  \caption{Prompt used to determine the injection points of the hallucinated answers to source texts for hallucination type scene.
  }
  \label{fig:det_inject_scene}
\end{figure*}

Figures \ref{fig:det_inject_att}, \ref{fig:det_inject_rel}, and \ref{fig:det_inject_scene} illustrate the prompts used to guide GPT-4 in determining injection points for attribute, relationship, and scene type questions.
Note that identifying injection points for object type questions is unnecessary, as it is evident that the paragraphs do not include the hallucinated object.

\subsection{Injecting Hallucinated Answers}
\label{sec:injecting_algorithm}
After identifying possible injection points, we prompt GPT-4 to inject the hallucinated answer to the paragraph. 

\begin{figure*}[t!]
  \centering
  \includegraphics[width=0.72\linewidth]{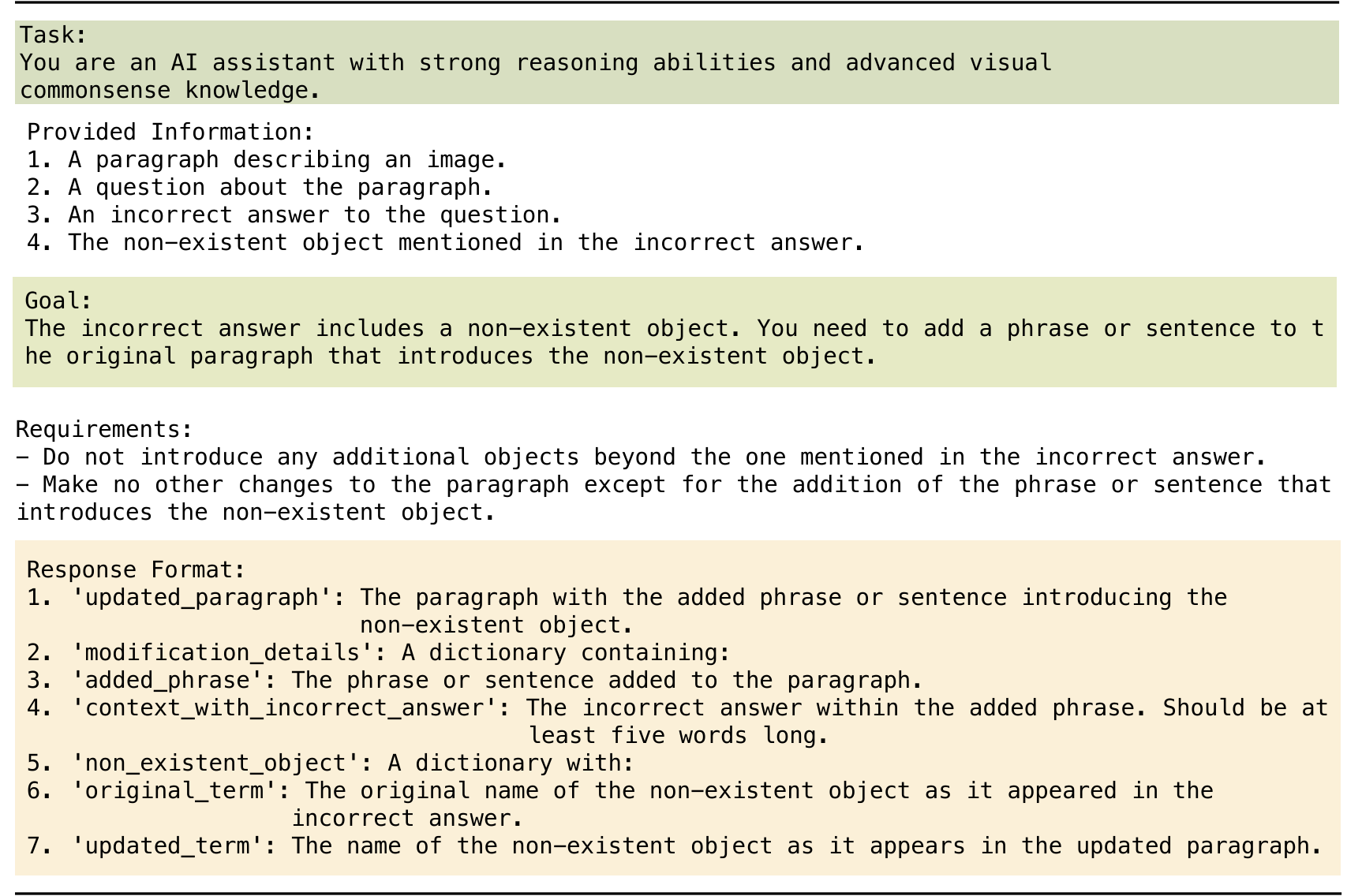}
  \caption{Prompt to inject hallucinated answers pertaining to \(\texttt{<obj>}\) hallucination.
  }
  \label{fig:injecting_object}
\end{figure*}

\begin{figure*}[t!]
  \centering
  \includegraphics[width=0.72\linewidth]{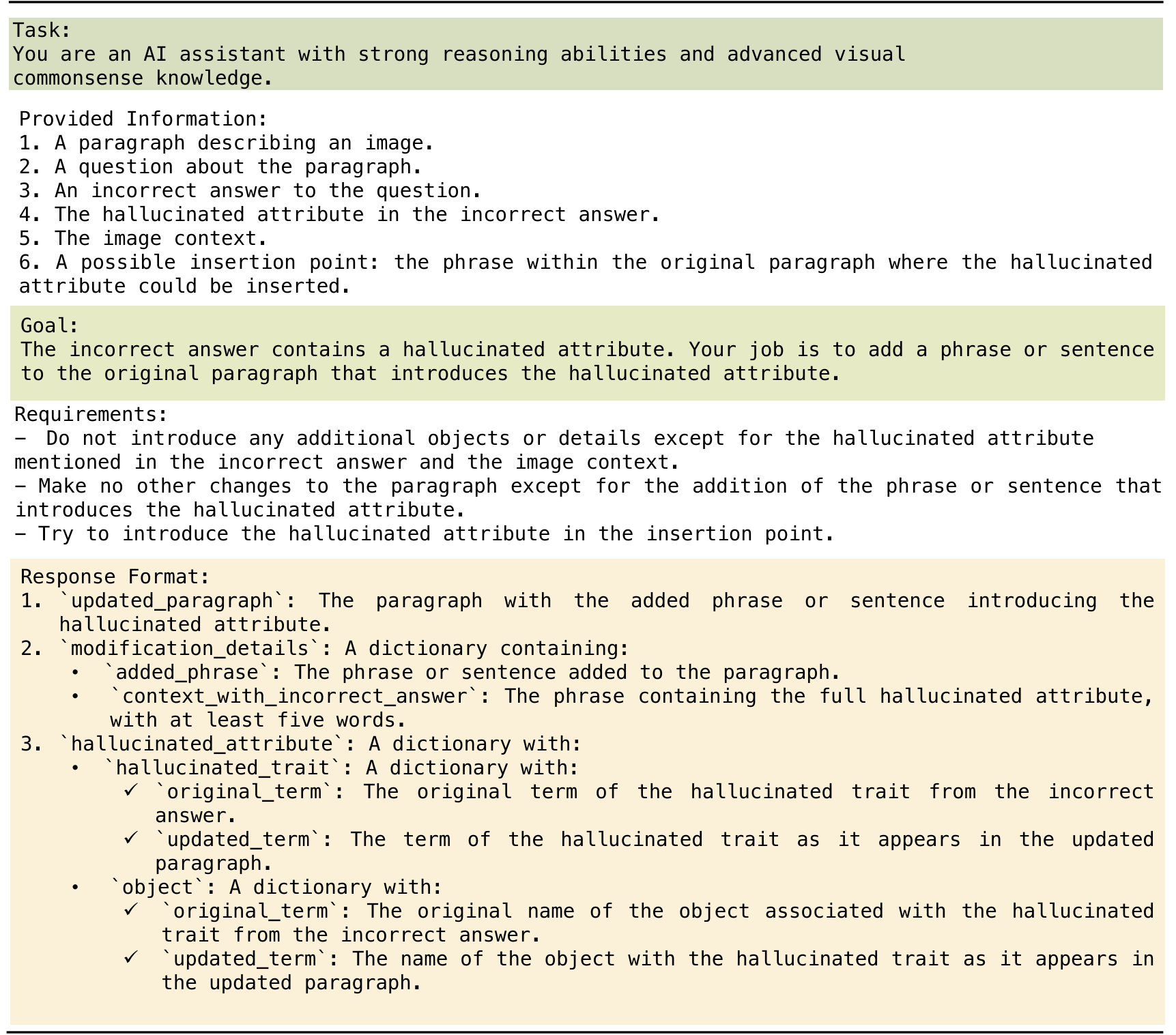}
  \caption{Prompt to inject hallucinated answers pertaining to \(\texttt{<attr>}\) hallucination.
  }
  \label{fig:injecting_attr}
\end{figure*}

\begin{figure*}[t!]
  \centering
  \includegraphics[width=0.7\linewidth]{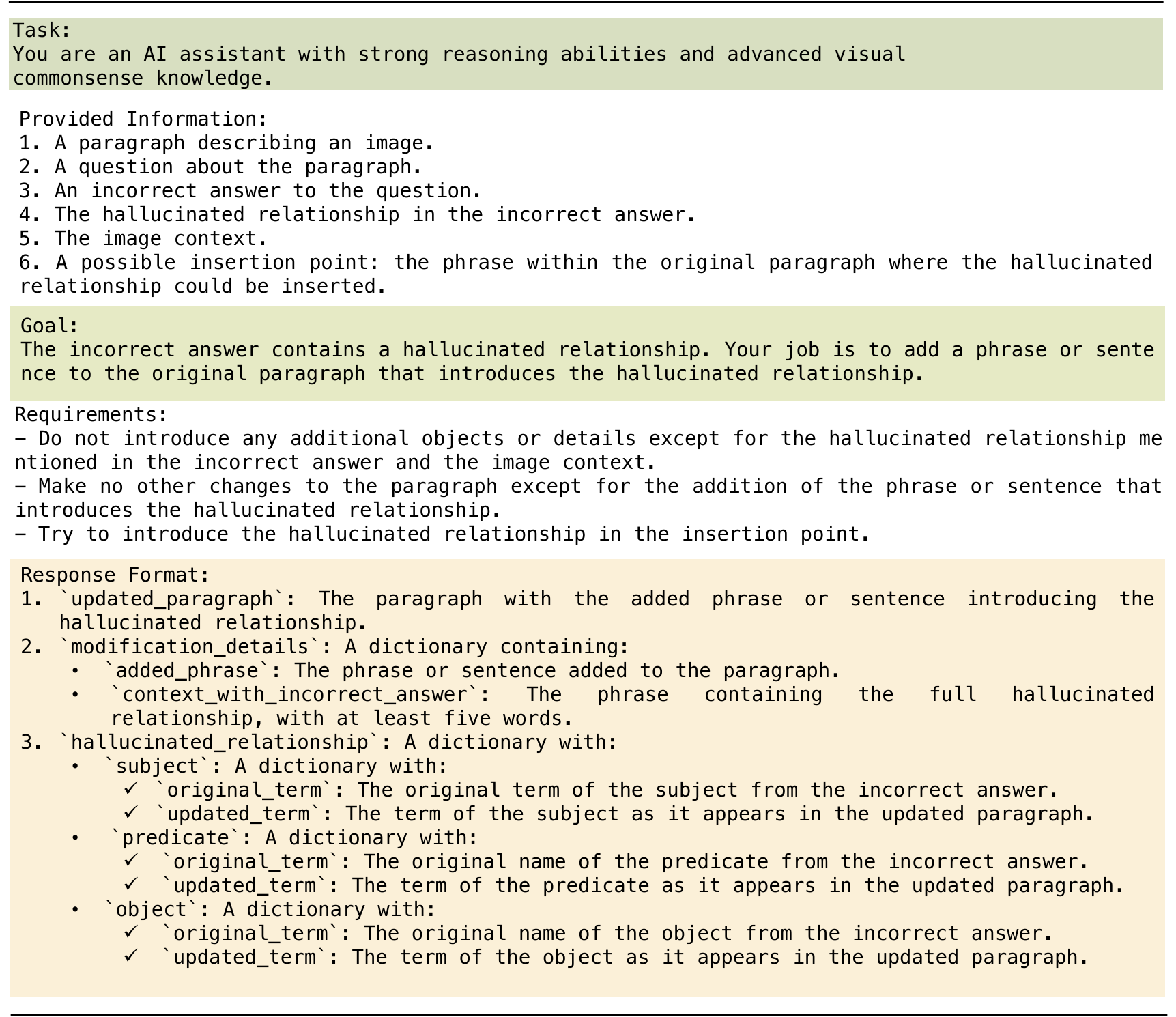}
  \caption{Prompt to inject hallucinated answers pertaining to \(\texttt{<rel>}\) hallucination.
  }
  \label{fig:injecting_rel}
\end{figure*}

\begin{figure*}[t!]
  \centering
  \includegraphics[width=0.7\linewidth]{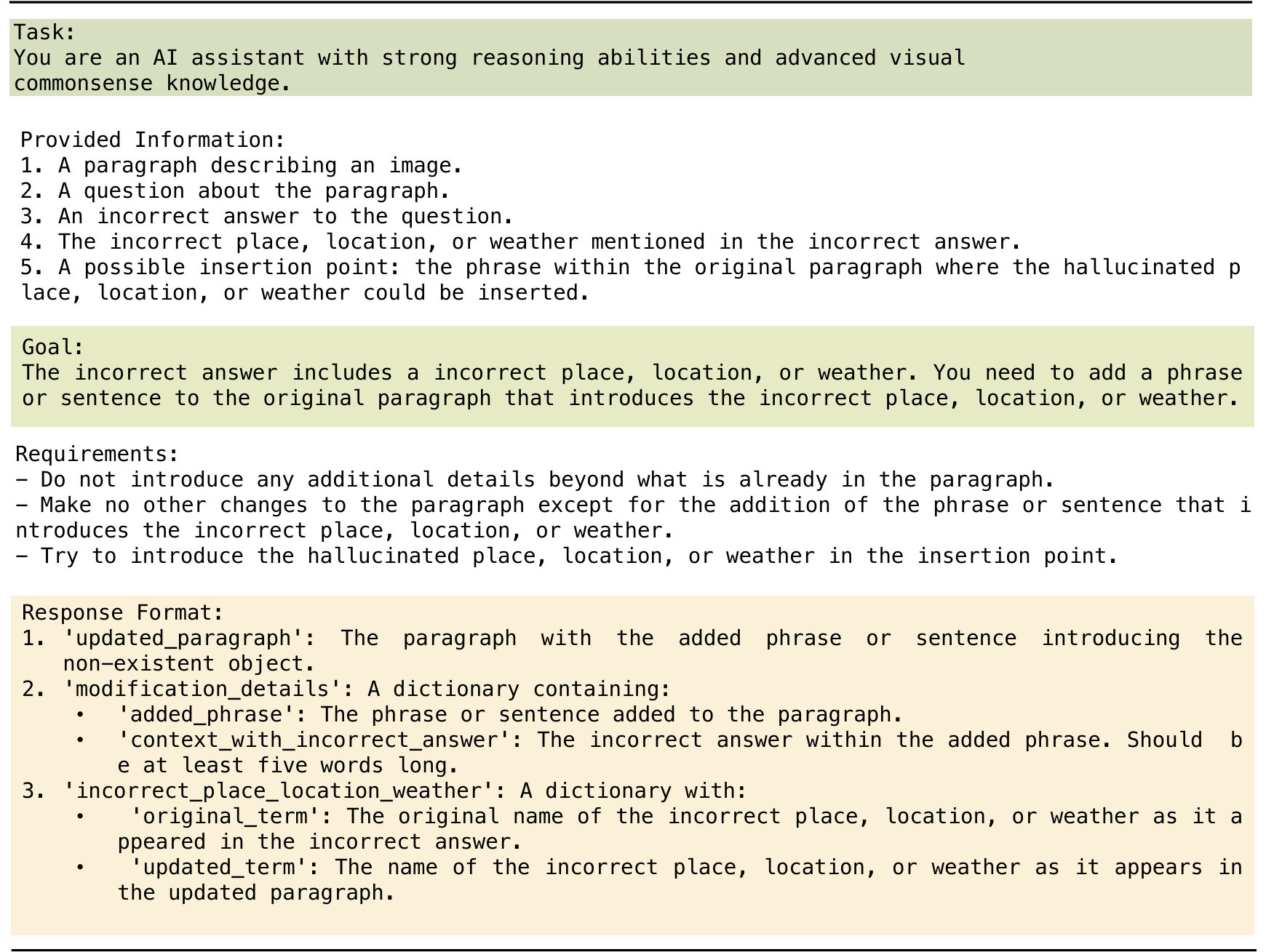}
  \caption{Prompt to inject hallucinated answers pertaining to \(\texttt{<sce>}\) hallucination.
  }
  \label{fig:injecting_scene}
\end{figure*}

Figures \ref{fig:injecting_object}, \ref{fig:injecting_attr}, \ref{fig:injecting_rel}, and \ref{fig:injecting_scene} illustrate the prompts used to inject hallucinated answers for object, attribute, relationship, and scene type questions, respectively.

\subsection{Verifying Injection}
After each injection, we verify whether the injection pipeline properly inserted the hallucinated answer.
This step is necessary because, despite specific instructions and rule-based filtering, we have observed several instances where GPT-4 fails to inject hallucinated answers correctly. In particular, while creating \mybenchmarkp, which requires multiple rounds of HQA injection, we observed success rates of 57\%, 33\%, 30\%, and 61\% for object, attribute, relationship, and scene questions, respectively.

\begin{figure*}[t!]
  \centering
  \includegraphics[width=0.72\linewidth]{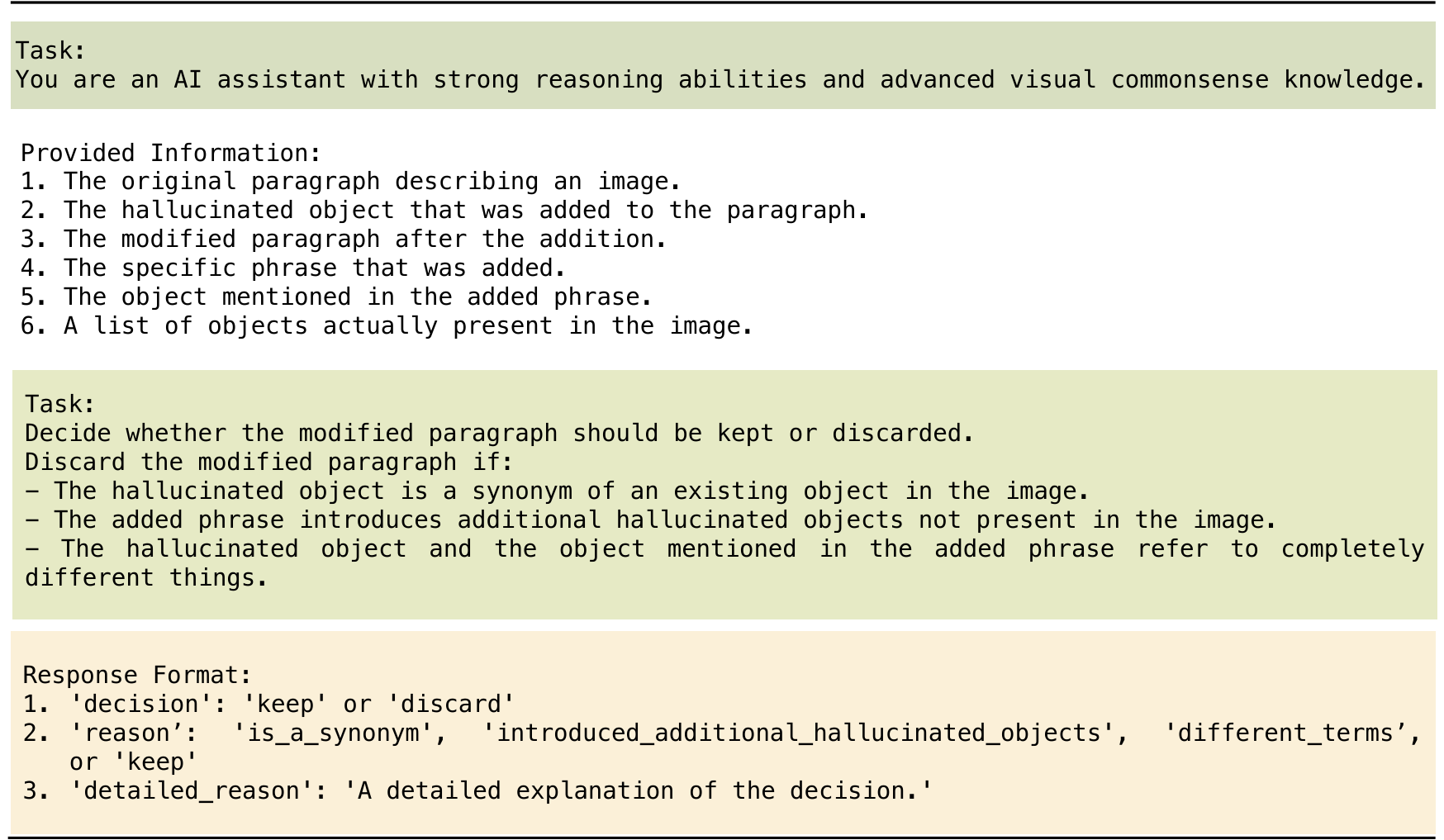}
  \caption{Prompt to verify each injection step of injecting hallucinated answers for \(\texttt{<obj>}\) hallucination.
  }
  \label{fig:verifying_obj}
\end{figure*}

\begin{figure*}[t!]
  \centering
  \includegraphics[width=0.72\linewidth]{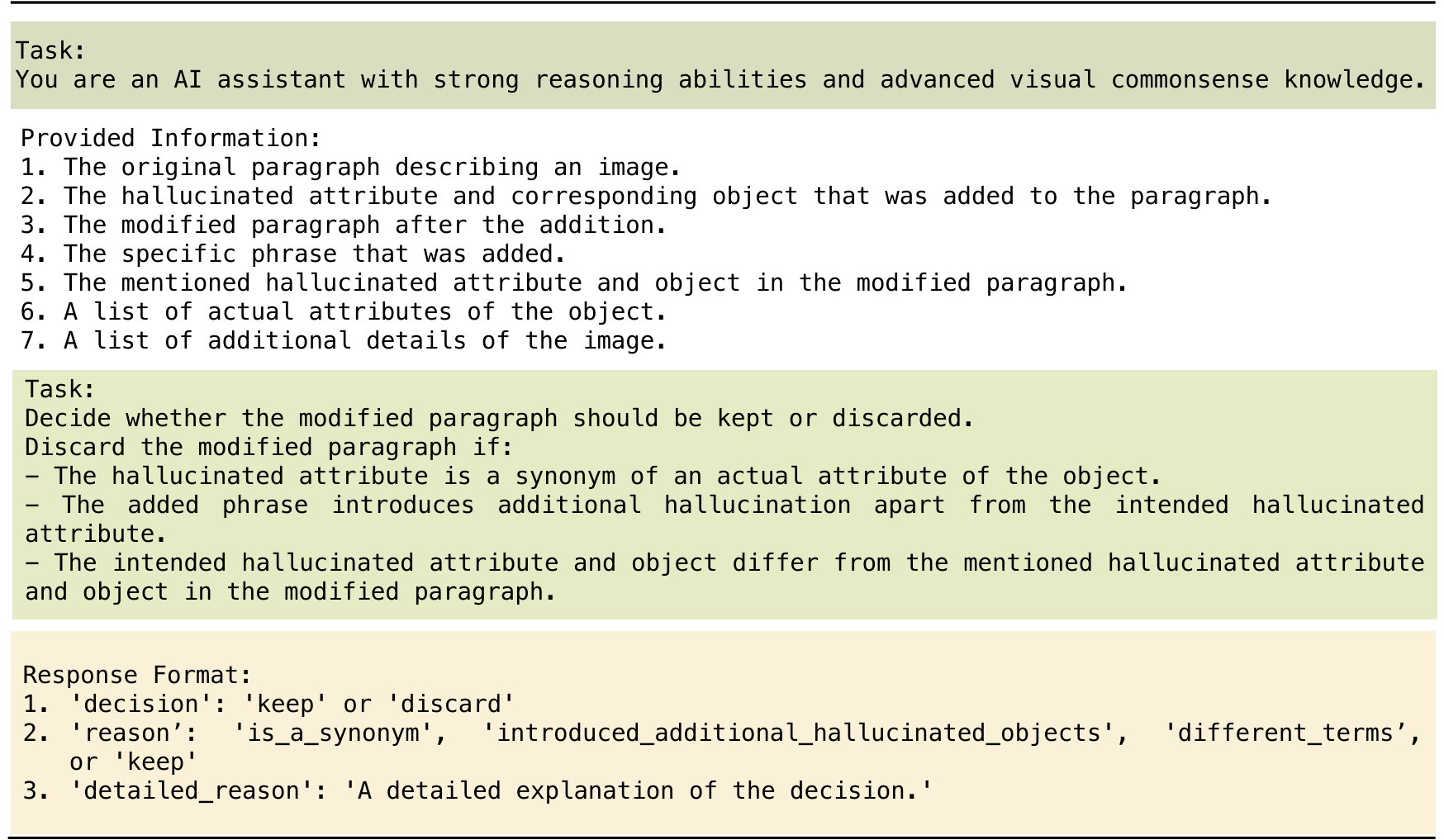}
  \caption{Prompt to verify each injection step of injecting hallucinated answers for \(\texttt{<attr>}\) hallucination.
  }
  \label{fig:verifying_attr}
\end{figure*}

\begin{figure*}[t!]
  \centering
  \includegraphics[width=0.7\linewidth]{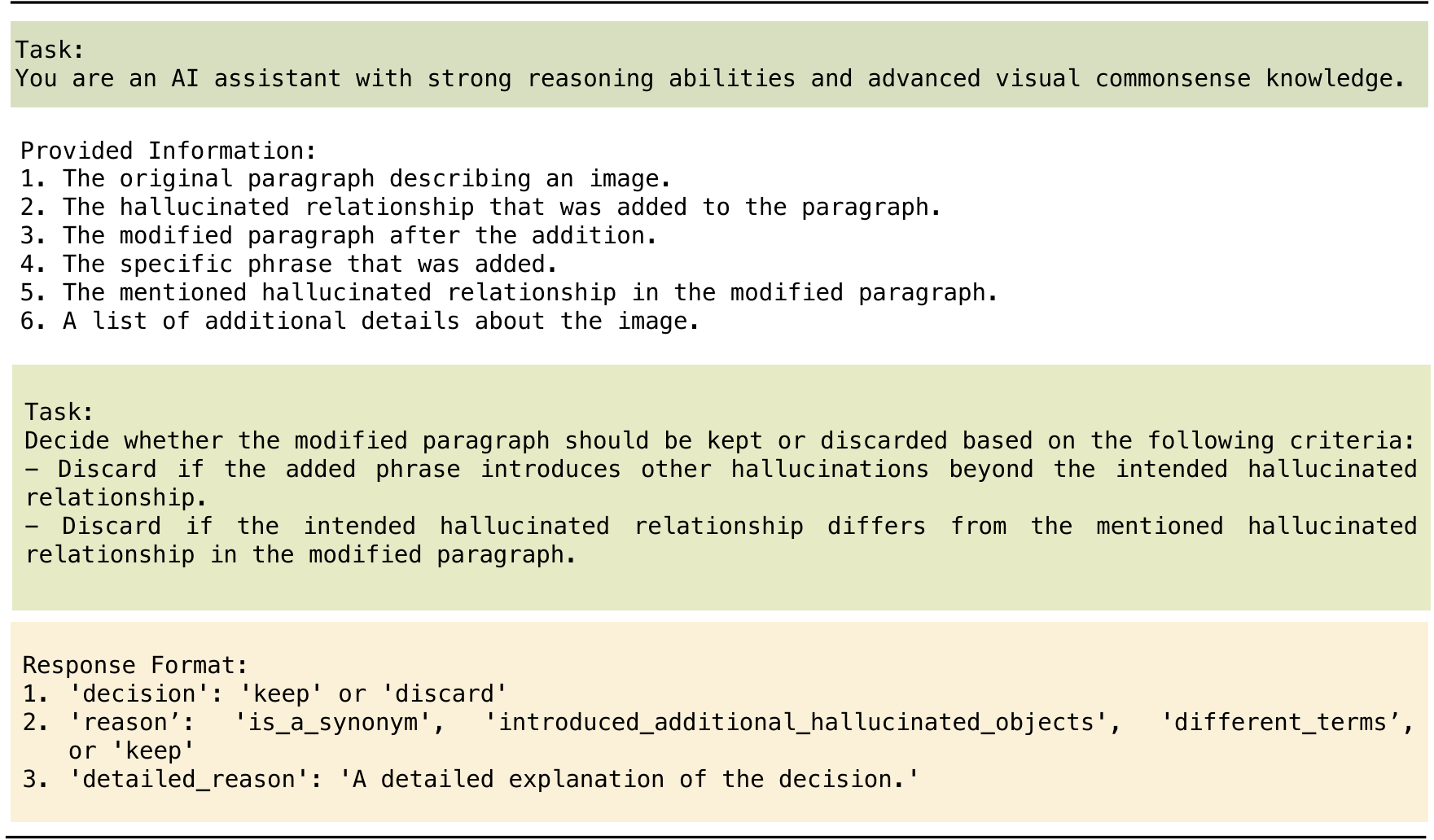}
  \caption{Prompt to verify each injection step of injecting hallucinated answers for \(\texttt{<rel>}\) hallucination.
  }
  \label{fig:verifying_rel}
\end{figure*}

\begin{figure*}[t!]
  \centering
  \includegraphics[width=0.7\linewidth]{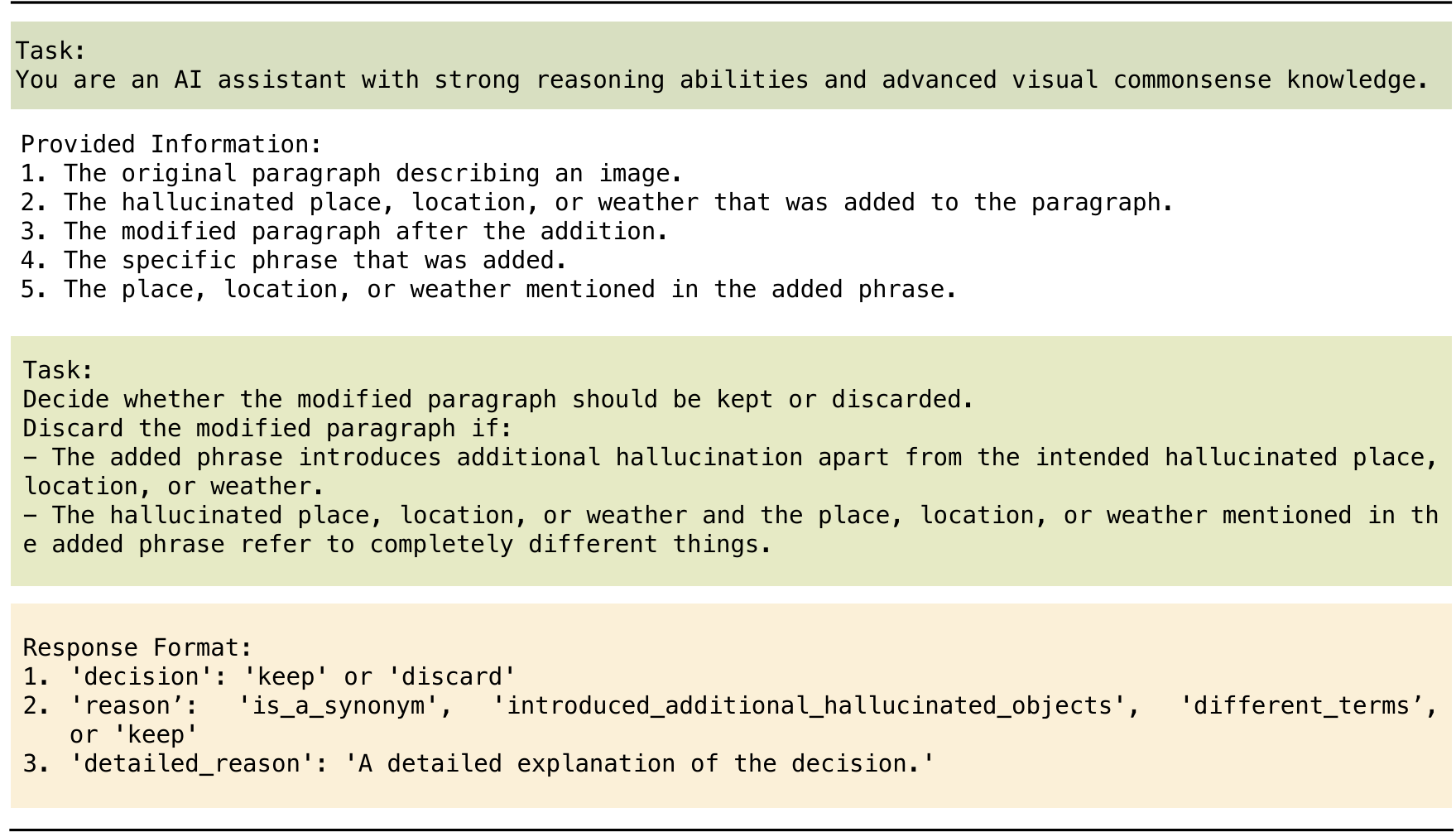}
  \caption{Prompt to verify each injection step of injecting hallucinated answers for \(\texttt{<sce>}\) hallucination.
  }
  \label{fig:verifying_scene}
\end{figure*}

Figures \ref{fig:verifying_obj} , \ref{fig:verifying_attr}, \ref{fig:verifying_rel}, and \ref{fig:verifying_scene} show the prompts used to verify each injection step for object, attribute, relationship, and scene questions. Refer to Algorithm \ref{qa_injection_algorithm} for a complete algorithmic overview of the HQA injection. 

\begin{algorithm}
\caption{HQA Injection}
\label{qa_injection_algorithm}
\begin{algorithmic}[1]
\REQUIRE Paragraph $P$, Number of HQA pairs $n$, QA database $Q$
\STATE Select $n$ HQA pairs from $Q$
\STATE Initialize $curr\_paragraph \gets P$
\STATE Initialize $coarse\_annotations \gets \emptyset$
\FOR{each question-hallucinated answer pair $i \in [1, n]$}
    \STATE Extract components $a, b, c$ from $i$
    \IF{answer $c$ in $curr\_paragraph$}
        \STATE $injection\_point \gets c$
    \ELSIF{component $a$ or $b$ in $curr\_paragraph$}
        \STATE $injection\_point \gets$ phrase surrounding $a$ or $b$
    \ELSE
        \STATE $injection\_point \gets$ selected by GPT-4
    \ENDIF

    \STATE Use GPT-4 to inject the hallucinated answer at the $injection\_point$
    \STATE Obtain $modified\_paragraph$, $phrase$, and $components$
    
    \IF{$modified\_paragraph$ does not contain phrases in $coarse\_annotations$}
        \STATE Continue to next question
    \ENDIF
    \IF{$phrase$ not unique in $modified\_paragraph$}
        \STATE Continue to next question
    \ENDIF
    \IF{$phrase$ does not contain components}
        \STATE Continue to next question
    \ENDIF
    
    \IF{additional hallucinations introduced}
        \STATE Continue to next question
    \ENDIF
    \IF{hallucinated answer differs from hallucination in $phrase$}
        \STATE Continue to next question
    \ENDIF
    
    \STATE Update $curr\_paragraph$ to $modified\_paragraph$
    \STATE Add $phrase$, $components$ to $coarse\_annotations$
\ENDFOR

\FOR{each $ann$ in $coarse\_annotations$}
    \STATE Find index of $phrase$ in $modified\_paragraph$
    \STATE Find index of $components$ in $phrase$
    \STATE Annotate each component with index
\ENDFOR

\end{algorithmic}
\end{algorithm}

\clearpage


\clearpage

\section{Qualitative Analysis of \mymodel}
\label{halloc_qualitative}
We provide a qualitative analysis of the performance of \mymodel\. We use the \mymodel\ trained with LLaVA \cite{liu2023visual} embeddings to conduct the analysis.

\begin{figure*}[t!]
  \centering
  \includegraphics[width=\linewidth]{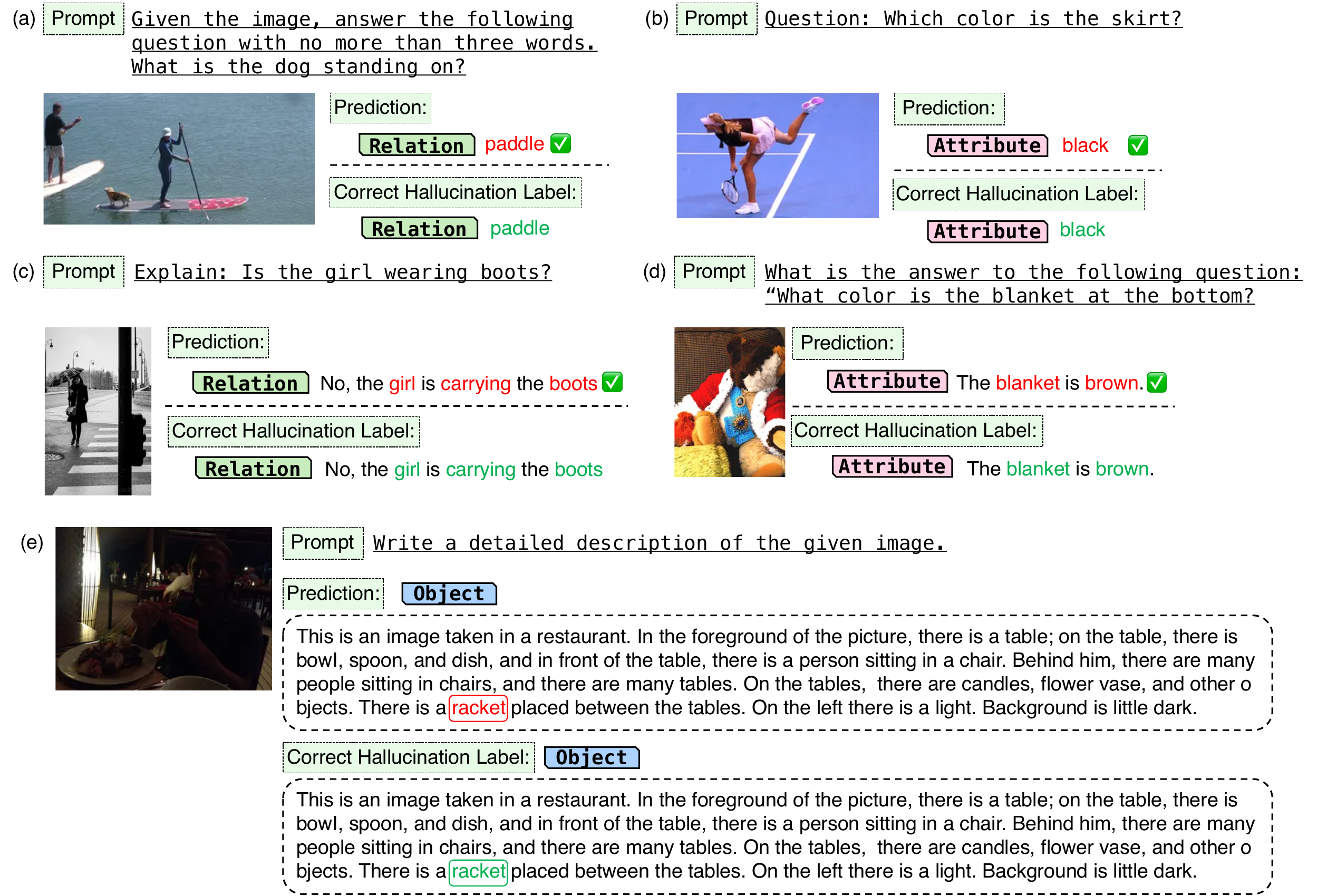}
  \caption{Success cases of HalLoc: Correctly localizes and identifies hallucination type (a) relationship and (b) attribute in \mybenchmarks. Correctly localizes and identifies hallucination type (c) relationship and (d) attribute in \mybenchmarki. Correctly localizes and identifies hallucination type (e) object in \mybenchmarkp. 
  }
  \label{fig:success_cases}
\end{figure*}

\begin{figure*}[t!]
  \centering
  \includegraphics[width=\linewidth]{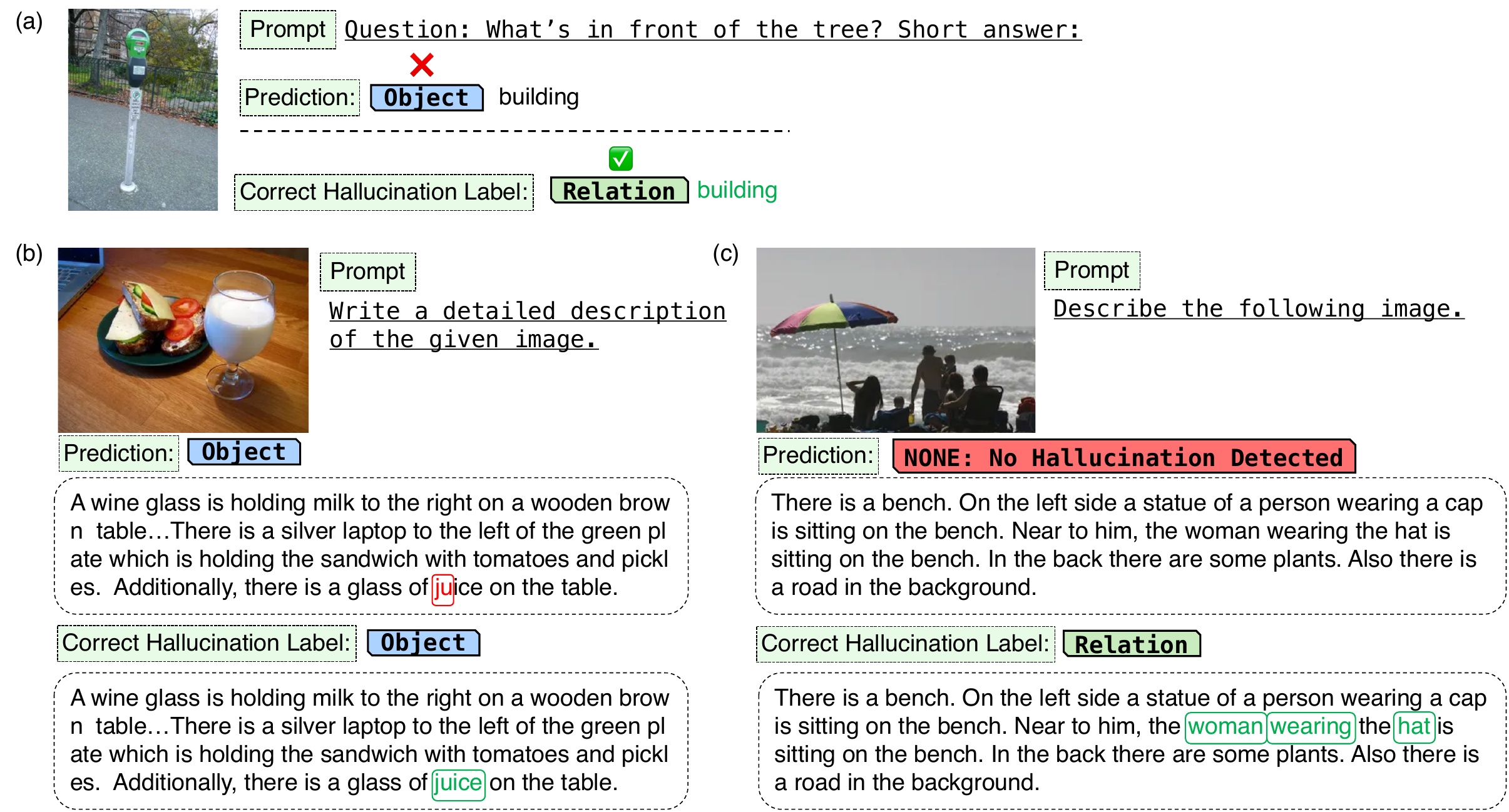}
  \caption{Failure cases: (a) Correctly detects hallucination but assigns the wrong  hallucination type. (b) Correctly assigns the hallucination type but incorrect token-level detection. (c) Case where no hallucination was detected when there is a relationship hallucination present in the hallucinated text.
  }
  \label{fig:failure_cases}
\end{figure*}

\begin{figure*}[t!]
  \centering
  \includegraphics[width=\linewidth]{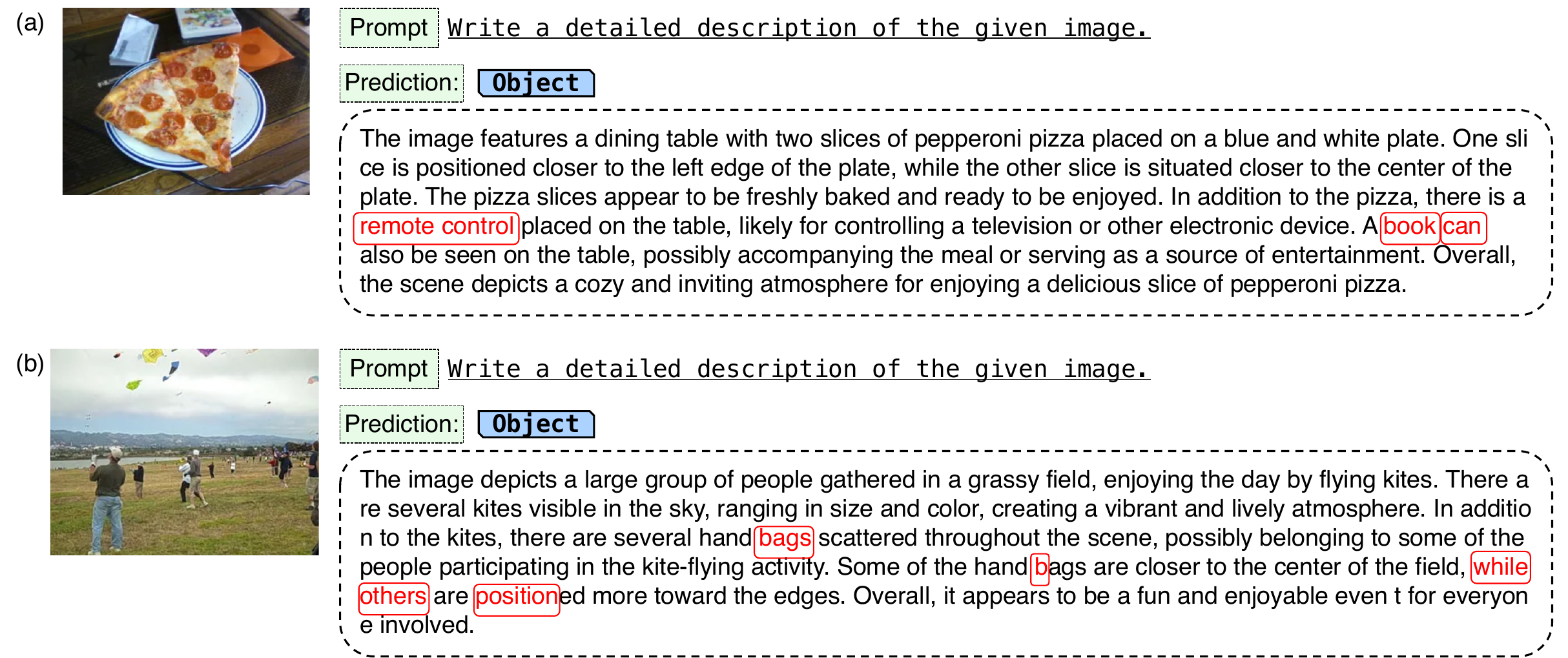}
  \caption{Examples where \mymodel\ detects and identifies hallucinations in image captions generated by a vision language model. We used LLaVA \cite{liu2023visual} for our example.
  }
  \label{fig:real_world_examples}
\end{figure*}

\subsection{Success Scenarios}
Figure \ref{fig:success_cases} presents examples where \mymodel\ successfully localized hallucinations and correctly identified their types. (a) and (b) in Figure \ref{fig:success_cases} show successful cases from \mybenchmarks, while (c) and (d) depict successful examples from \mybenchmarki. Figure \ref{fig:success_cases} (e) illustrates a successful case from \mybenchmarkp.

\subsection{Failure Scenarios}
Figure \ref{fig:failure_cases} illustrates several instances where \mymodel\ did not perform as expected on the test cases from \mybenchmark.

(a) in Figure \ref{fig:failure_cases} depicts a frequent failure pattern where \mymodel\ incorrectly identifies a non-existent object in the image as a relationship type. This error occurs because the question prompts for a relationship, but \mybenchmark\ classifies any non-existent object as an object type rather than a relationship.

(b) in Figure \ref{fig:failure_cases} presents another common failure, where \mymodel\ only partially identifies components of hallucination. Such failure cases show that \mymodel\ is imperfect in identifying word boundaries, which should be explored further in future research. 

(c) in Figure \ref{fig:failure_cases} highlights a recurring issue, particularly with \mybenchmarkp, where \mymodel\ fails to detect any hallucinated tokens. This problem likely stems from the sparse presence of hallucinated tokens in lengthy responses. Future research should explore innovative approaches to address this challenge.

\subsection{\mymodel\ in Real-World Scenarios}
We evaluate \mymodel\ on free-form responses generated by real-world VLMs to assess its performance in practical settings beyond \mybenchmark. Figure \ref{fig:real_world_examples} showcases case studies where \mymodel\ is applied to detailed descriptions generated by LLaVA \cite{liu2023visual}. These examples highlight the robustness and adaptability of \mymodel\ when dealing with the complexity and variability of real-world data.

\clearpage

{
   \small
   \bibliographystyle{ieeenat_fullname}
   \bibliography{main}
}
\end{document}